\documentclass[journal]{IEEEtran}
\ifCLASSINFOpdf
\else
\fi

\usepackage{booktabs}
\usepackage{tabularx}
\usepackage{multirow}
\usepackage{graphics}
\usepackage{amsmath}
\usepackage{mathrsfs}
\usepackage{amssymb}
\usepackage{pifont}
\usepackage{tcolorbox}
\usepackage{url}
 
\makeatletter
\def\@seccntformat#1{\csname the#1\endcsname.\quad}

\def\@IEEEsectpunct{.\ } % section 标题后句点+空格
\def\@IEEEsectpunctspace{ }

\renewcommand\subsubsection{\@startsection{subsubsection}{3}{\z@}%
  {0.8\baselineskip}% 上方间距
  {0.3\baselineskip}% 下方间距
  {\normalfont\normalsize\itshape}} % 保留 IEEE italic 样式

\def\@IEEEsectpunct{.\par}
\makeatother

\begin{document}
%
% paper title
% Titles are generally capitalized except for words such as a, an, and, as,
% at, but, by, for, in, nor, of, on, or, the, to and up, which are usually
% not capitalized unless they are the first or last word of the title.
% Linebreaks \\ can be used within to get better formatting as desired.
% Do not put math or special symbols in the title.
\title{SafePLUG: Empowering Multimodal LLMs with Pixel-Level Insight and Temporal Grounding for Traffic Accident Understanding}
%
%
% author names and IEEE memberships
% note positions of commas and nonbreaking spaces ( ~ ) LaTeX will not break
% a structure at a ~ so this keeps an author's name from being broken across
% two lines.
% use \thanks{} to gain access to the first footnote area
% a separate \thanks must be used for each paragraph as LaTeX2e's \thanks
% was not built to handle multiple paragraphs
%

\author{
Zihao Sheng$^{\rm 1}$, Zilin Huang$^{\rm 1}$, Yansong Qu$^{\rm 2}$, Jiancong Chen$^{\rm 2}$, Yuhao Luo$^{{\rm 1}}$, \\Yen-Jung Chen$^{\rm 3}$, Yue Leng$^{{\rm 4}}$, Sikai Chen$^{\rm 1, *}$% <-this % stops a space
\thanks{$^{\rm 1}$ Department of Civil and Environmental Engineering, University of Wisconsin-Madison, Madison, WI 53706, USA}% <-this % stops a space
\thanks{$^{\rm 2}$ Lyles School of Civil and Construction Engineering, Purdue University, West Lafayette, IN 47907, USA}% <-this % stops a space
\thanks{$^{\rm 3}$ Elmore Family School of Electrical and Computer Engineering, Purdue University, West Lafayette, IN 47907, USA}
\thanks{$^{\rm 4}$ Google, Sunnyvale, CA 94089, USA}
\thanks{$^{\rm *}$ Corresponding author: sikai.chen@wisc.edu}
}

% note the % following the last \IEEEmembership and also \thanks - 
% these prevent an unwanted space from occurring between the last author name
% and the end of the author line. i.e., if you had this:
% 
% \author{....lastname \thanks{...} \thanks{...} }
%                     ^------------^------------^----Do not want these spaces!
%
% a space would be appended to the last name and could cause every name on that
% line to be shifted left slightly. This is one of those "LaTeX things". For
% instance, "\textbf{A} \textbf{B}" will typeset as "A B" not "AB". To get
% "AB" then you have to do: "\textbf{A}\textbf{B}"
% \thanks is no different in this regard, so shield the last } of each \thanks
% that ends a line with a % and do not let a space in before the next \thanks.
% Spaces after \IEEEmembership other than the last one are OK (and needed) as
% you are supposed to have spaces between the names. For what it is worth,
% this is a minor point as most people would not even notice if the said evil
% space somehow managed to creep in.

% The paper headers
\markboth{Journal of \LaTeX\ Class Files}%
{IEEEtran.cls for IEEE Journals}
% The only time the second header will appear is for the odd numbered pages
% after the title page when using the twoside option.
% 
% *** Note that you probably will NOT want to include the author's ***
% *** name in the headers of peer review papers.                   ***
% You can use \ifCLASSOPTIONpeerreview for conditional compilation here if
% you desire.

% If you want to put a publisher's ID mark on the page you can do it like
% this:
%\IEEEpubid{0000--0000/00\$00.00~\copyright~2015 IEEE}
% Remember, if you use this you must call \IEEEpubidadjcol in the second
% column for its text to clear the IEEEpubid mark.

% use for special paper notices
%\IEEEspecialpapernotice{(Invited Paper)}

% make the title area
\maketitle

% As a general rule, do not put math, special symbols or citations
% in the abstract or keywords.
\begin{abstract}
Multimodal large language models (MLLMs) have achieved remarkable progress across a range of vision-language tasks and demonstrate strong potential for traffic accident understanding. However, existing MLLMs in this domain primarily focus on coarse-grained image-level or video-level comprehension and often struggle to handle fine-grained visual details or localized scene components, limiting their applicability in complex accident scenarios. To address these limitations, we propose \textbf{SafePLUG}, a novel framework that empowers MLLMs with both \textbf{P}ixel-\textbf{L}evel \textbf{U}nderstanding and temporal \textbf{G}rounding for comprehensive traffic accident analysis. SafePLUG supports both arbitrary-shaped visual prompts for region-aware question answering and pixel-level segmentation based on language instructions, while also enabling the recognition of temporally anchored events in traffic accident scenarios. To advance the development of MLLMs for traffic accident understanding, we curate a new dataset, \textit{SafePLUG-Bench}, which contains diverse multimodal question–answer pairs with detailed pixel-level annotations and temporal event boundaries across a wide range of accident scenarios. 
Experimental results show that SafePLUG achieves strong performance on multiple tasks, including region-based question answering, pixel-level segmentation, temporal event localization, and accident event understanding. These capabilities lay a foundation for fine-grained understanding of complex traffic scenes, with the potential to improve driving safety and enhance situational awareness in smart transportation systems.
Code, data, and models will be publicly available at \url{https://zihaosheng.github.io/SafePLUG/}.
\end{abstract}

% Note that keywords are not normally used for peerreview papers.

% For peer review papers, you can put extra information on the cover
% page as needed:
% \ifCLASSOPTIONpeerreview
% \begin{center} \bfseries EDICS Category: 3-BBND \end{center}
% \fi
%
% For peerreview papers, this IEEEtran command inserts a page break and
% creates the second title. It will be ignored for other modes.
\IEEEpeerreviewmaketitle

\section{Introduction}

Recent advances in multimodal large language models (MLLMs) have demonstrated remarkable capabilities in understanding and reasoning over visual and linguistic information, enabling a wide range of applications from visual question answering (QA) to video analysis~\cite{yin2024survey,caffagni2024revolution,liu2024improved}. Building on these successes, researchers have increasingly explored the potential of MLLMs within intelligent transportation systems and autonomous driving~\cite{li2024steering,huang2025vlm,sheng2025talk2traffic}, particularly in advancing traffic accident understanding~\cite{parikh2025roadsocial,xing2025echotraffic,zhou2025tau}. By jointly processing information across multiple modalities, MLLMs offer a promising paradigm for analyzing traffic incidents and answering complex queries~\cite{karim2025large,yan2025large,zhang2025language}. These capabilities can be valuable in a variety of real-world traffic scenarios. For example, drivers may benefit from real-time accident interpretation and warning feedback, while analysts and planners can use them to assist in post-accident review, liability assessment, and identifying common failure patterns~\cite{xu2021sutd,fang2024abductive,huang2025sky}. 

Understanding traffic accidents often requires fine-grained, pixel-level comprehension to ensure accurate identification of critical objects, spatial relationships, and impact regions. However, existing MLLMs in this domain~\cite{parikh2025roadsocial,xing2025echotraffic} primarily operate at a coarse granularity, focusing on global scene understanding at the image or video level while lacking the ability to localize and reason about specific regions involved in an accident. This coarse granularity hinders their ability to capture nuanced visual cues that are essential for accurate accident interpretation. In contrast, pixel-level MLLMs are capable of processing fine-grained visual details~\cite{ren2024pixellm,yan2024visa,rasheed2024glamm,munasinghe2025videoglamm}, with the potential to support more accurate segmentation of collision areas, detection of minor yet critical objects, and differentiation between overlapping agents. Furthermore, by leveraging arbitrary-shaped pixel-level visual prompts as input, the model can be better guided to attend to semantically and contextually relevant areas, enhancing its ability to filter out irrelevant background and improving accuracy on region-sensitive tasks~\cite{cai2024vip,lin2024draw,zhang2024omg}.

\begin{figure*}[t]
\centering
\includegraphics[width=0.99\textwidth]{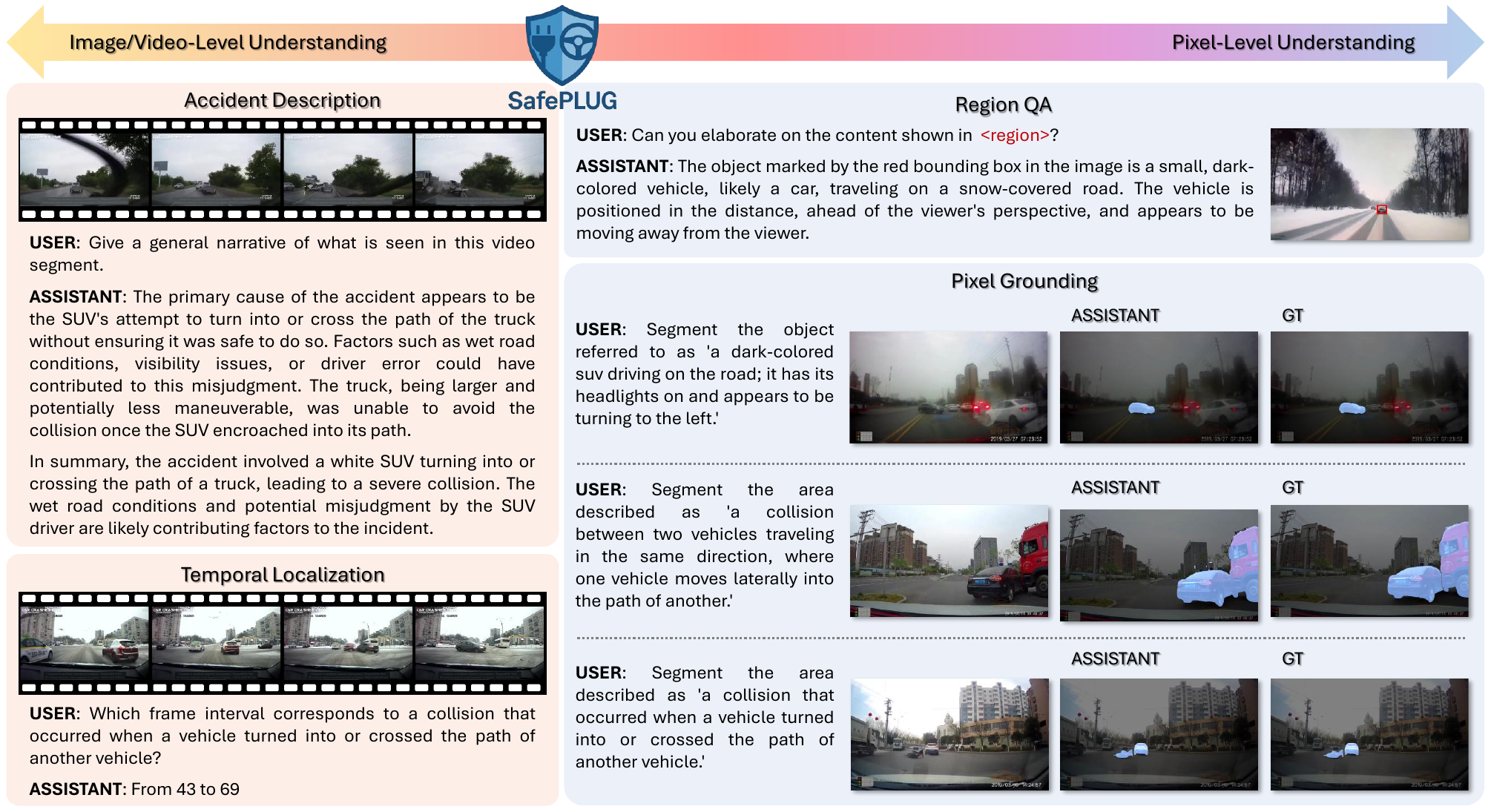}
\caption{SafePLUG supports both image/video-level and pixel-level understanding through accident description, temporal localization, region-level question answering, and pixel-level grounding, enabling comprehensive traffic accident analysis.}
\label{fig1}
\end{figure*}

Another critical aspect of traffic accident understanding is temporal grounding, which refers to identifying the start and end times of specific events within a video. In traffic accident understanding, knowing exactly when the accident occurs is essential for supporting fine-grained accident phase analysis. By distinguishing between pre-, during-, and post-accident phases, the model can separate normal driving behavior from abnormal actions, thereby enabling effective warnings~\cite{yao2022dota,fang2024abductive}. While recent video-based MLLMs have made substantial progress in recognizing what happens in a scene, they often struggle to determine when it happens~\cite{huang2024vtimellm,ren2024timechat,guo2025vtg}. This limitation arises because most models are trained to align visual content with language, focusing on semantic understanding rather than temporal localization~\cite{wu2025number}. The RoadSocial benchmark~\cite{parikh2025roadsocial} also highlights this gap by evaluating models on predicting event boundaries in traffic videos, and finds that even strong MLLMs often produce implausible time spans. These findings underscore the importance of equipping MLLMs with robust temporal grounding abilities to ensure reliable accident interpretation.

To bridge these gaps and advance the application of MLLMs in traffic accident understanding, we propose SafePLUG, a novel framework that empowers MLLMs with both \textbf{P}ixel-\textbf{L}evel \textbf{U}nderstanding and temporal \textbf{G}rounding capabilities. For pixel-level understanding, SafePLUG incorporates a visual prompt encoder that extracts region-aware features from arbitrary-shaped visual prompts and aligns them with the language instructions. We further extend the LLM vocabulary with a special \texttt{<SEG>} token, whose hidden embedding is utilized by a SAM-based decoder~\cite{kirillov2023segment} to produce pixel-wise segmentation masks. For temporal grounding, we incorporate a lightweight number prompt mechanism, in which unique numeric indicators are overlaid on video frames to implicitly convey temporal positions. By treating these numbers as visual cues, the model is guided to associate semantic events with specific temporal segments. Importantly, number prompts integrate seamlessly into the video input without modifying the model architecture or requiring additional training objectives. 
As illustrated in Figure~\ref{fig1}, SafePLUG exhibits remarkable capabilities across four key tasks: accident description, temporal localization, region-level QA, and pixel-level grounding. To support the development and evaluation of such models, we construct a new benchmark dataset containing multimodal question-answer pairs, pixel-wise annotations, and frame-level event boundaries across diverse accident scenarios.

In summary, our contributions are as follows:
\begin{itemize}
\item We propose SafePLUG, a novel framework that equips MLLMs with both pixel-level understanding and temporal grounding capabilities, enabling fine-grained reasoning over complex traffic accident scenarios through the integration of visual and number prompts.

\item We curate SafePLUG-Bench, a new benchmark dataset for traffic accident understanding. To the best of our knowledge, it is the first dataset in this domain that supports both region-level QA and pixel-level grounding QA.

\item Extensive experiments across multiple tasks, including region-level QA,  pixel-level segmentation, temporal event localization, and accident event understanding, demonstrate the superior performance of SafePLUG. All code, dataset, and model checkpoints will be released to facilitate future research.
\end{itemize}

\section{Related Work}

\subsection{Traffic Accident Understanding Methods}
Traffic accident understanding involves identifying key agents, detecting anomalies, and interpreting causal and temporal dynamics in complex driving scenarios~\cite{yao2022dota,xu2024tad,ling2024pedestrian,you2020traffic}. Earlier approaches primarily used CNN-based models to classify accidents or detect behavioral phases from visual inputs~\cite{santhosh2021vehicular,zhou2022spatio,fang2023vision,yin2025human}. However, these models lack high-level semantic reasoning and cannot answer open-ended questions, such as “Analyze why the accident happened.”

\begin{table*}[t]
\centering
\caption{Comparison of existing traffic accident understanding datasets with ours. Bbox: Bounding Box, TG: Temporal Grounding, PGQA: Pixel-level Grounding QA.}
\begin{tabular}{l | c | c | ccc | c | c | c}
\toprule  
\multirow{2}{*}{Dataset} & \multirow{2}{*}{\shortstack{Year}} & \multirow{2}{*}{\shortstack{Frames}} & \multicolumn{3}{c|}{Annotations} & \multirow{2}{*}{Region QA} & \multirow{2}{*}{PGQA} & \multirow{2}{*}{QA Pairs}  \\ 
\cmidrule(lr){4-6}
 &  &  & Bbox & Mask & TG &  &  &   \\
\midrule
A3D~\cite{yao2019unsupervised} & 2019 & 208K & \ding{51} & -- & \ding{51} & -- & -- & -- \\ 
CCD~\cite{bao2020uncertainty} & 2020 & 75K & \ding{51} & -- & \ding{51} & -- & -- & -- \\ 
DADA~\cite{fang2021dada}  & 2021 & 658K & \ding{51} & -- & \ding{51} & -- & -- & -- \\ 
DADA-Seg~\cite{zhang2021exploring}  & 2021 & 12K & -- & \ding{51} & \ding{51} & -- & -- & -- \\ 
SUTD-TrafficQA~\cite{xu2021sutd}  & 2021 & 1.90M & -- & -- & \ding{51} & -- & -- & 62K \\ 
DoTA~\cite{yao2022dota} & 2022 & 732K & \ding{51} & -- & \ding{51} & -- & -- & -- \\ 
MM-AU~\cite{fang2024abductive} & 2024 & 2.19M & \ding{51} & -- & \ding{51} & -- & -- & 58K \\ 
TAU-106K~\cite{zhou2025tau} & 2025 & -- & \ding{51} & -- & \ding{51} & -- & -- & 332K \\ 
RoadSocial~\cite{parikh2025roadsocial} & 2025 & 14M & -- & -- & \ding{51} & -- & -- & 260K \\ 
AV-TAU~\cite{xing2025echotraffic} & 2025 & 3.16M & -- & -- & \ding{51} & -- & -- & 149K \\ 
SafePLUG-Bench (Ours)  & 2025 & 2.26M & \ding{51} & \ding{51} & \ding{51} & \ding{51} & \ding{51} & 220K \\ 
\bottomrule
\end{tabular}
\label{tab:dataset}
\end{table*}

Recently, MLLMs have been introduced to enhance traffic accident understanding. EchoTraffic~\cite{xing2025echotraffic} incorporates audio cues to improve the anomaly reasoning capabilities of MLLMs. \cite{parikh2025roadsocial} demonstrate that fine-tuning general video MLLMs on their proposed dataset improves model performance in road event comprehension. TABot~\cite{zhou2025tau} combines functional and instruction tuning for MLLMs, and leverages bounding box information to provide spatial grounding of accident regions and involved agents. \cite{guan2025domain} further introduce a domain-enhanced dual-branch framework that integrates multimodal features through large models such as GPT-4o and Long-CLIP. Their model jointly learns visual–temporal representations and domain knowledge embeddings, enabling efficient and interpretable accident anticipation.

Nonetheless, existing models lack pixel-level visual understanding and rely solely on refined instruction datasets with annotated timestamps for temporal localization. Their spatial reasoning is limited to bounding boxes, without support for segmentation or region-level QA. In contrast, our work extends MLLMs to support diverse input and output modalities, including visual prompts, number prompts, and segmentation masks, enabling fine-grained spatial reasoning and temporally grounded accident analysis.

\subsection{Traffic Accident Understanding Datasets}

Early traffic accident datasets are primarily designed to support tasks such as accident detection, accident type classification, and identification of involved objects~\cite{chan2016anticipating,lv2021localizing}. The A3D dataset~\cite{yao2019unsupervised} provides annotations for accident categories, bounding boxes of involved objects, and timestamps indicating when accidents are identified. DoTA~\cite{yao2022dota} extends A3D by incorporating more videos and richer annotations, including anomaly types, related objects, and tracking IDs. The CCD dataset~\cite{bao2020uncertainty} further offers accident causes for each video sequence, while DADA~\cite{fang2021dada} explores the role of driver attention in traffic accident prediction by collecting eye-gaze data. Based on this, DADA-Seg~\cite{zhang2021exploring} refines a subset of 313 video sequences with fine-grained segmentation masks for semantic objects. Although these datasets have significantly advanced visual-based traffic accident analysis, they primarily support coarse-grained tasks and lack detailed language annotations.

In recent years, an increasing number of datasets have emerged to support traffic accident understanding through language-based QA. SUTD-TrafficQA~\cite{xu2021sutd} is the first large-scale benchmark in this domain, offering six types of video-QA pairs, such as accident description, forecasting, and reasoning. MM-AU~\cite{fang2024abductive} provides textual annotations that cover three key aspects of traffic accidents: causality, prevention strategies, and accident types. TAU-106K~\cite{zhou2025tau} advances this direction with questions requiring temporal localization and spatial grounding, where textual answers include timestamps and bounding box coordinates. The RoadSocial dataset~\cite{parikh2025roadsocial} further broadens the task scope with diverse video QAs for general road events. Meanwhile, AV-TAU~\cite{xing2025echotraffic} enriches the multimodal context of traffic accident scenarios by incorporating audio signals.

Our dataset, SafePLUG-Bench, further advances the field by uniquely supporting both region-level QA and pixel-level grounding QA. It is densely annotated with segmentation masks and includes over 220K high-quality multimodal QA pairs across diverse accident scenarios. A comprehensive comparison with existing datasets is summarized in Table~\ref{tab:dataset}.

\section{Method}
\begin{figure*}[t]
\centering
\includegraphics[width=0.99\textwidth]{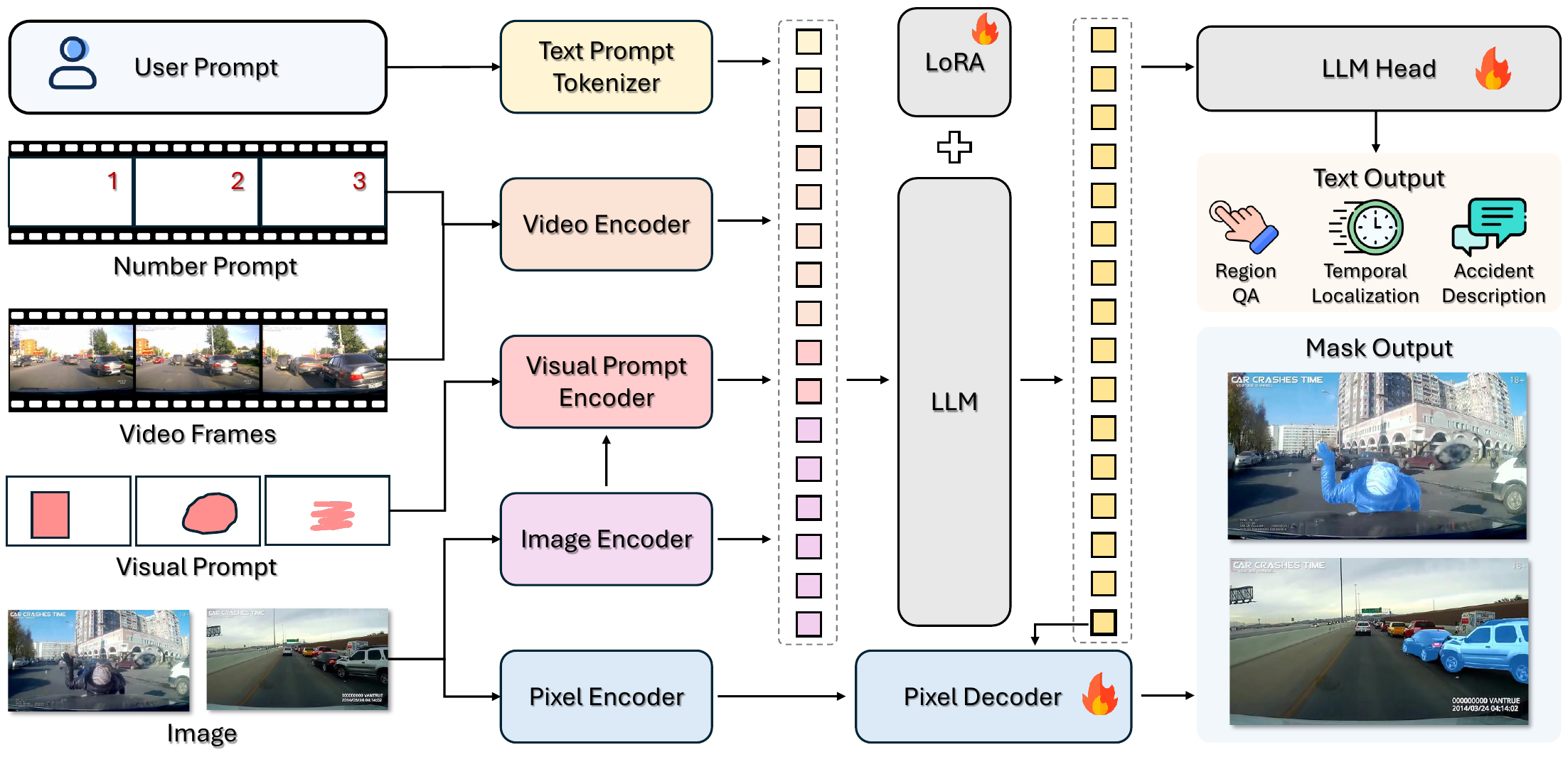}
\caption{Overview of SafePLUG. The model takes as input multiple modalities, including video frames with number prompts, image-level context, and user-defined visual prompts, and unifies them with language prompts through an LLM backbone. The features are then decoded into either natural language answers or binary segmentation masks.}
\label{fig2}
\end{figure*}

Figure~\ref{fig2} illustrates the overall architecture of the proposed SafePLUG framework, which extends MLLMs with enhanced pixel-level understanding and temporal grounding abilities for comprehensive traffic accident analysis. Unlike prior approaches that focus primarily on global scene comprehension, SafePLUG introduces fine-grained spatial reasoning and temporally aligned perception to capture subtle accident dynamics. In the following subsections, we elaborate on each module in detail.

\subsection{Multimodal Input Encoding}
\subsubsection{Video Encoder with Number Prompts}
To represent the temporal evolution of traffic scenarios, we denote a video sequence as $V \in \mathbb{R}^{T\times H\times W\times 3}$, where $T$ is the number of uniformly sampled frames from the raw footage. The entire sequence $V$ is fed into a pretrained video encoder $\mathcal{E}_{\text{vid}}$ from LanguageBind~\cite{zhu2023languagebind} to extract spatiotemporal representations:
\begin{equation}
\mathbf{V} = \mathcal{E}_{\text{vid}}(V), \quad \mathbf{V} \in \mathbb{R}^{T \times L_v \times d_v},
\end{equation}
where $L_v=H' \times W'$ denotes the number of tokens per frame, and $d_v$ is the hidden dimension of the video encoder.
The resulting features $\mathbf{V}$ are then transformed by a projection layer $\mathbf{h}_{\text{vid}} = \mathcal{P}_{\text{vid}}(\mathbf{V})$ to align with the input space of the language model.
To further strengthen the model’s temporal grounding capability, we introduce a lightweight yet effective enhancement, i.e., number prompts~\cite{wu2025number}. During preprocessing, a numerical indicator $n_t$ corresponding to the frame index is visually overlaid onto each frame, providing cues for the model to identify the temporal order of events. The numerical markers are strategically placed (e.g., top-right corner) to preserve visual content integrity. This simple augmentation does not alter the model architecture or objective but has been shown to guide the model in accurately associating semantic events with their temporal boundaries.

\subsubsection{Visual Prompt Encoder}
To enable fine-grained spatial reasoning within complex traffic scenes, SafePLUG incorporates a visual prompt encoder that encodes user-defined regions of interest (ROIs) such as polygons or arbitrary-shaped masks. Given an image $I \in \mathbb{R}^{H\times W\times 3}$ and a corresponding binary mask $M \in \{0,1\}^{H \times W}$ that specifies the visual prompt region, we extract intermediate feature maps $\mathbf{F} \in \mathbb{R}^{H' \times W' \times d_i}$ from the pretrained image encoder $\mathcal{E}_{\text{img}}$. The region-aware feature representation is then obtained through masked average pooling:
\begin{equation}
\mathbf{r} = \frac{1}{\|M'\|_1} \sum_{x,y} M'_{x,y}\,\mathbf{F}_{x,y},
\end{equation}
where $M'$ denotes the downsampled version of $M$ aligned with $\mathbf{F}$. The embedding $\mathbf{r} \in \mathbb{R}^{d_i}$ is projected into the LLM input space through a learnable projection layer $\mathcal{P}_{\text{roi}}$. 
This representation encodes localized visual semantics conditioned on the prompt region, allowing the model to selectively attend to contextually relevant areas. By combining explicit spatial prompts with language cues, the visual prompt encoder provides a flexible and effective mechanism for grounding linguistic instructions in specific image regions.

\subsubsection{Image and Pixel Encoder}
To capture both holistic scene context and fine-grained spatial cues, SafePLUG integrates two complementary visual encoders.  
First, a pretrained image encoder $\mathcal{E}_{\text{img}}$ extracts visual features from the image $I$:
\begin{equation}
\mathbf{i} = \mathcal{E}_{\text{img}}(I), \quad \mathbf{i} \in \mathbb{R}^{L_i \times d_i}.
\end{equation}
These high-level embeddings encode the global semantics of the scene and serve as the contextual foundation for downstream reasoning tasks such as region-level question answering.
% In parallel, SafePLUG employs a pixel-level encoder $\mathcal{E}_{\text{pix}}$ built upon the Segment Anything Model (SAM)~\cite{kirillov2023segment}, which extracts dense spatial features $\mathbf{P} \in \mathbb{R}^{H_p \times W_p \times d_p}$ suitable for fine-grained segmentation. These pixel embeddings are later utilized by the segmentation decoder to produce fine-grained masks aligned with language instructions. Together, the image and pixel encoders enable SafePLUG to reason over global semantics and local spatial details, ensuring both contextual understanding and precise localization in complex traffic scenes.

Second, a pixel-level encoder $\mathcal{E}_{\text{pix}}$ built upon the Segment Anything Model (SAM)~\cite{kirillov2023segment}
extracts dense spatial features from $I$:
\begin{equation}
\mathbf{p} = \mathcal{E}_{\text{pix}}(I), \quad \mathbf{p} \in \mathbb{R}^{H_p \times W_p \times d_p},
\end{equation}
where $\mathbf{p}$ denotes the pixel-level embeddings capturing fine-grained spatial cues.
These features are subsequently utilized by the SAM-based decoder $\mathcal{D}_{\text{pix}}$ to generate segmentation masks aligned with language instructions.
Together, the image encoder $\mathcal{E}_{\text{img}}$ and pixel encoder $\mathcal{E}_{\text{pix}}$ enable SafePLUG to reason over global semantics and local spatial details, ensuring both contextual understanding and precise localization in complex traffic scenes.

\subsection{Multimodal Fusion via LLM}
SafePLUG integrates multimodal representations within a unified large language model to enable joint spatial–temporal reasoning. 
The visual features extracted from the video, image, and region encoders are each projected into the language embedding space through their respective projectors $\mathcal{P}_{\text{vid}}$, $\mathcal{P}_{\text{img}}$, and $\mathcal{P}_{\text{roi}}$. 
During tokenization, we insert special placeholder tokens (\texttt{<video>}, \texttt{<image>}, and \texttt{<region>}) into the input text sequence. The projected visual embeddings then replace these placeholders to form a multimodal input sequence:
\begin{equation}
\mathbf{X} = T_{\text{text}} 
\big|_{\texttt{<video>} \rightarrow \mathcal{P}_{\text{vid}}(\mathbf{V}),\;
       \texttt{<image>} \rightarrow \mathcal{P}_{\text{img}}(\mathbf{i}),\;
       \texttt{<region>} \rightarrow \mathcal{P}_{\text{roi}}(\mathbf{r})}.
\end{equation}
The interleaved sequence $\mathbf{X}$ is then processed by the language model $f_{\rm llm}$, which jointly attends to textual and visual information for cross-modal reasoning across spatial regions, temporal sequences, and linguistic context. This fusion design allows SafePLUG to achieve coherent, context-aware interpretation of traffic scenes from multiple modalities.

\subsection{Decoders for Textual and Pixel Output} 
SafePLUG produces two types of outputs (i.e., natural language responses and pixel-level segmentation masks) within a unified decoding framework. 
For textual tasks, the language model directly generates responses through its language head, producing token probabilities conditioned on the multimodal input sequence. 
For pixel-level tasks, following prior works~\cite{lai2024lisa,huang2025towards,yuan2025sa2va}, we extend the LLM vocabulary with a special token \texttt{<SEG>}, whose hidden representation $\mathbf{h}_{\texttt{<SEG>}}$ serves as a query vector for segmentation. After processing the multimodal input, $\mathbf{h}_{\texttt{<SEG>}}$ is transformed by a learnable projector $\mathcal{P}_{\text{seg}}$ to form a segmentation prompt, which is jointly provided with the dense spatial features $\mathbf{p}$ extracted by the SAM pixel encoder $\mathcal{E}_{\text{pix}}$ as inputs to the SAM decoder $\mathcal{D}_{\text{pix}}$. The decoder then produces the final binary mask aligned with the textual instruction. This design allows SafePLUG to seamlessly switch between generating descriptive explanations and grounded visual outputs.

\subsection{Training Strategy}
SafePLUG adopts a modular training strategy inspired by the mixture-of-experts (MoE) paradigm~\cite{jacobs1991adaptive,fedus2022switch,sheng2024kinematics}, where two specialized sets of LoRA parameters~\cite{hu2022lora} are optimized to handle distinct output modalities. Rather than relying on a single adaptation pathway for all tasks, SafePLUG maintains a shared multimodal backbone and two lightweight, task-specific LoRA branches: one dedicated to natural language generation and the other to pixel-level segmentation. This design allows each branch to specialize in its respective modality while preserving shared visual–linguistic representations through a unified backbone.

The first LoRA branch focuses on text-oriented tasks, including accident description, region-level QA, and temporal localization. During this phase, the LoRA adapters are integrated throughout both the attention and feed-forward layers of the language model to enable parameter-efficient fine-tuning across the entire transformer structure. A small subset of projection and normalization layers connecting visual and textual representations is also softly updated to enhance cross-modal alignment. The model is optimized using a standard cross-entropy loss for text generation:
\begin{equation}
\mathcal{L}_{\text{text}} = - \sum_{i=1}^{T} \log P(y_i \mid y_{<i}, \mathbf{X}),
\end{equation}
where $y_i$ denotes the $i$-th target token and $\mathbf{X}$ represents the multimodal input sequence. 

The second branch is tailored for mask generation and spatial grounding. Building upon the pretrained backbone from the language branch, LoRA adapters are selectively applied only to the feed-forward layers to refine spatial reasoning while keeping the attention structure intact. In parallel, a compact set of segmentation-specific modules that map hidden language features to dense pixel embeddings are fine-tuned. The optimization objective combines binary cross-entropy (BCE) and DICE losses to balance pixel coverage accuracy and boundary precision:
\begin{equation}
\mathcal{L}_{\text{seg}} = \lambda_{\text{BCE}} \, \mathcal{L}_{\text{BCE}} + \lambda_{\text{DICE}} \, \mathcal{L}_{\text{DICE}},
\end{equation}
where $\lambda_{\text{BCE}}$ and $\lambda_{\text{DICE}}$ are weighting coefficients to balance pixel coverage and boundary precision.

After training, the two LoRA branches can be independently loaded or jointly combined depending on the target task. The language-oriented branch governs open-ended textual reasoning and explanation generation, whereas the mask-oriented branch provides pixel-level grounding and segmentation capabilities. Since both branches share the same multimodal encoders and tokenization pipeline, SafePLUG can dynamically activate the corresponding LoRA module according to the task type, enabling seamless switching between linguistic and spatial reasoning without performance interference or retraining.

\begin{figure}[t]
\centering
\includegraphics[width=0.9\columnwidth]{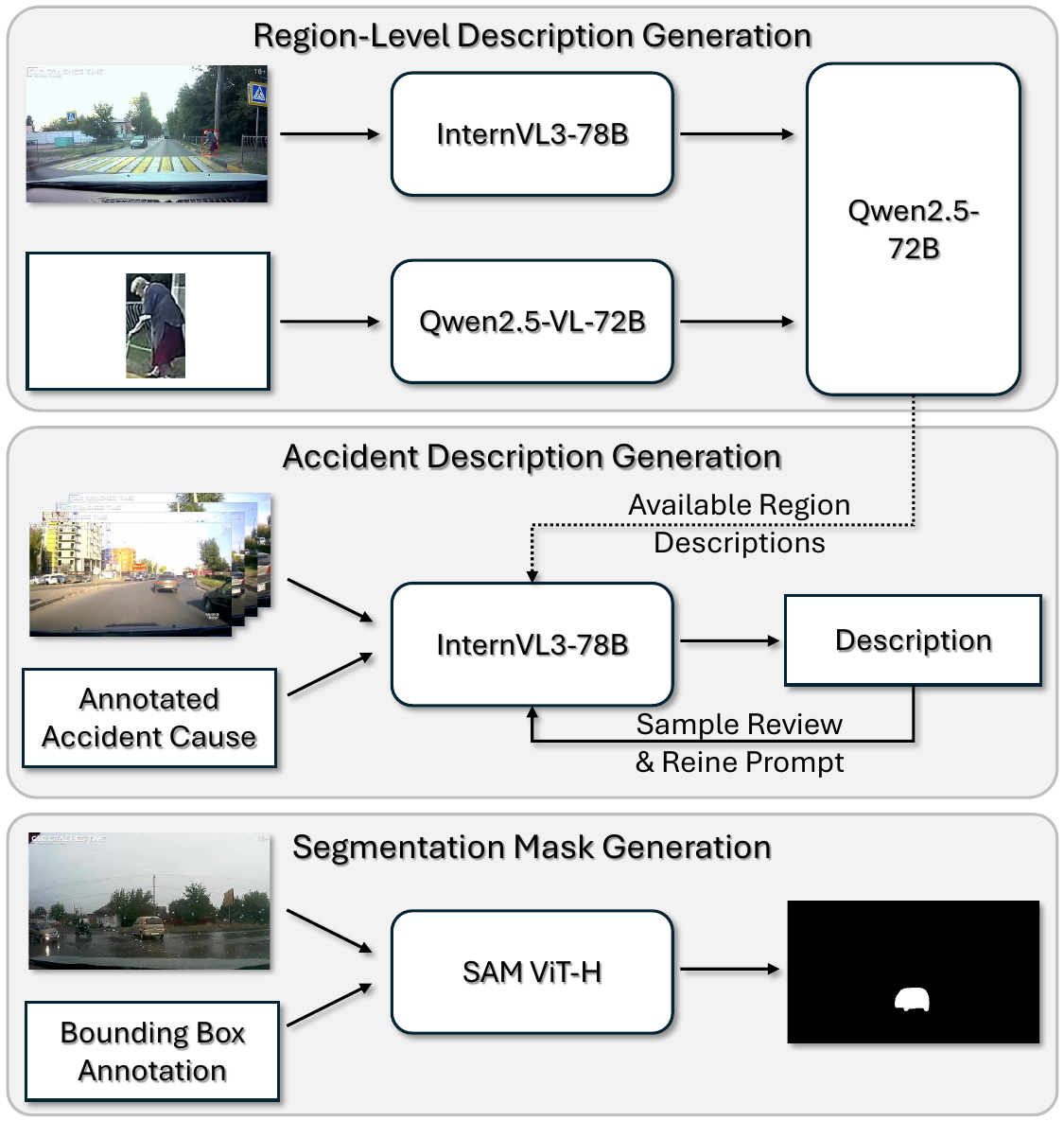}
\caption{Semi-automated data annotation pipeline leveraging MLLMs and SAM for generating region-level descriptions, accident narratives, and segmentation masks.}
\label{fig3}
\end{figure}

\begin{figure*}[t]
\centering
\includegraphics[width=0.9\textwidth]{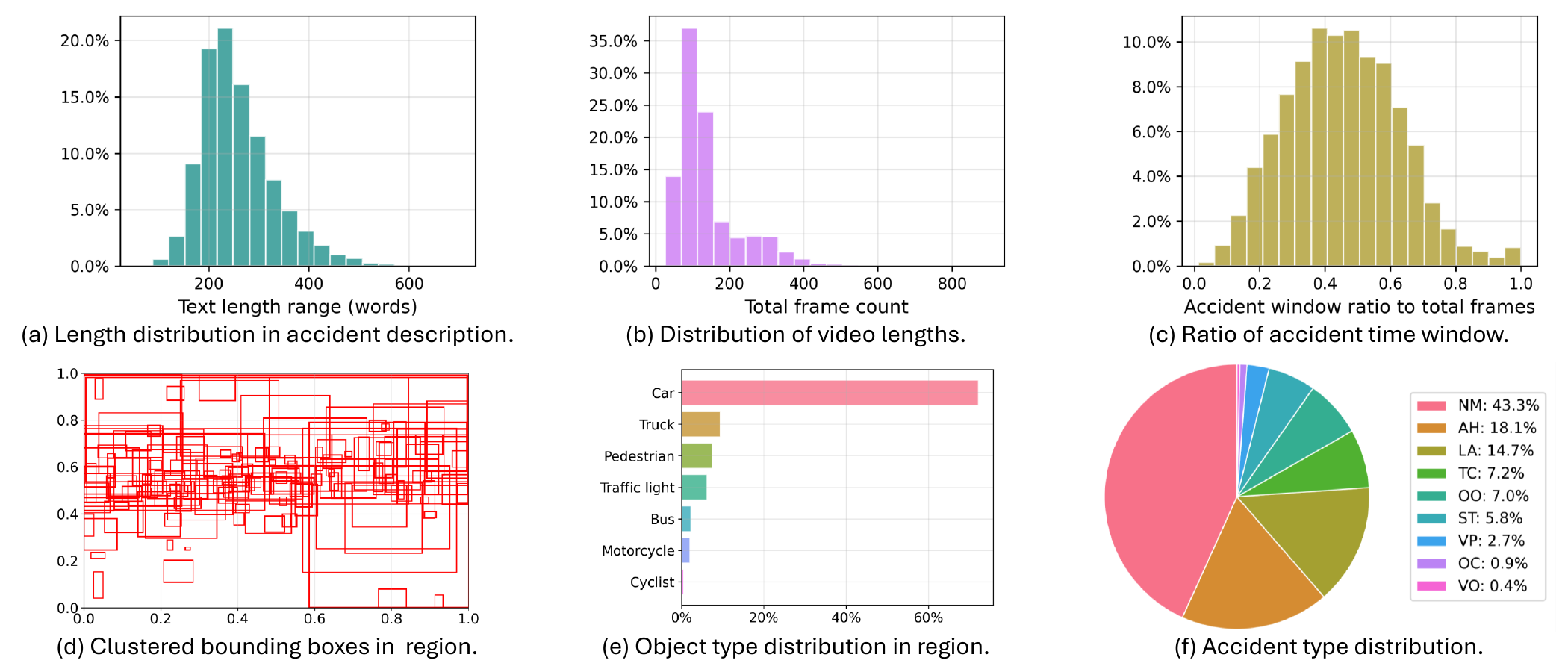}
\caption{Data statistics of SafePLUG-Bench, including text length, video duration, accident time ratio, spatial region density, object distribution, and accident type distribution.}
\label{fig3-2}
\end{figure*}

\subsection{SafePLUG-Bench}

Existing datasets for traffic accident understanding primarily focus on high-level visual QA but often lack fine-grained annotations required for tasks such as region-level QA and pixel-level grounding QA. These two capabilities are essential for enabling models to reason about specific regions involved in an incident and to localize accident-related semantics at the pixel level. To bridge this gap, we construct SafePLUG-Bench, a new benchmark dataset that supports both region QA and pixel-level grounding QA, in addition to standard accident description and temporal grounding tasks.

\subsubsection{Dataset Construction}
SafePLUG-Bench builds upon two existing benchmarks: DoTA~\cite{yao2022dota} and MM-AU~\cite{fang2024abductive}. As illustrated in Figure~\ref{fig3}, we adopt a semi-automated annotation pipeline that combines state-of-the-art multimodal foundation models with expert human verification to ensure both scalability and annotation fidelity. 

For region-level QA, we utilize the bounding box annotations from the original datasets. For each region, we generate two independent candidate descriptions: one by feeding the full image with overlaid bounding boxes to InternVL3-78B~\cite{chen2024internvl}, and another by inputting the cropped region to Qwen2.5-VL-72B~\cite{qwen2.5-VL}. The resulting descriptions are then cross-verified by Qwen2.5-72B~\cite{qwen2.5} to ensure semantic consistency, and inconsistent pairs are discarded. This process produces region-level answers that emphasize spatial context and local object relationships.

For accident description, we employ InternVL3-78B~\cite{chen2024internvl} to generate comprehensive narrative explanations of traffic incidents. Each generation process integrates visual information from video frames with auxiliary textual cues, i.e., annotated accident causes extracted from the original datasets. For DoTA, which provides detailed bounding boxes of involved agents, we further incorporate their verified region descriptions as additional contextual inputs. This multimodal prompting design allows the model to reason over both visual and causal dimensions, producing more coherent and causally grounded accident narratives. To ensure reliability and linguistic consistency, a subset of generated outputs was manually examined, based on which we iteratively refined the input prompt before finalizing the full dataset annotation pipeline.

For temporal localization, we treat each annotated accident cause as a textual query and its corresponding timestamp as the answer, establishing explicit temporal boundaries for each event. For pixel-level grounding QA, we employ SAM to automatically generate segmentation masks based on bounding box annotations, followed by human filtering to remove coarse or fragmented masks. Each valid region is then paired with its verified region description as the question and the corresponding mask as the answer. To enable holistic reasoning, all accident-related masks within a single frame are also merged into an aggregated ``accident-region” mask that identifies the full spatial extent of the incident. Additional details on the data annotation pipeline, prompt design, and human verification procedures are provided in Appendix~\ref{sec:appendix_dataset}.

\subsubsection{Dataset Statistics}

In total, SafePLUG-Bench comprises over 220K high-quality multimodal QA pairs covering diverse real-world accident scenarios. To ensure fair evaluation across all tasks, we randomly sample 500 QA pairs for each of the four subtasks. Figure~\ref{fig3-2} summarizes the overall statistical characteristics of the dataset.

Figure~\ref{fig3-2}(a) shows the word length distribution of answers in the Accident Description task. Most responses contain between 150 and 300 words. This indicates that the generated descriptions provide sufficient detail to capture causal reasoning and contextual information.
Figure~\ref{fig3-2}(b) illustrates the distribution of video lengths in frames. Most clips contain between 80 and 160 frames, providing moderate-length driving scenes that include both normal and accident phases. The ratio of the annotated accident window to the total video duration is shown in Figure~\ref{fig3-2}(c). The majority of accident windows occupy 30–60\% of the full sequence, such that each sequence includes adequate pre- and post-event context.

\begin{table*}[t]
\centering
\caption{Performance comparison on region QA and pixel grounding. All metric scores range from 0 to 100, with the best performance highlighted in \textbf{bold}.}
\begin{tabular}{l |c| cccc | cccc}
\toprule
\multirow{2}{*}{\textbf{Model}} & \multirow{2}{*}{\textbf{Param.}} & \multicolumn{4}{c|}{\textbf{Region QA}} & \multicolumn{4}{c}{\textbf{Pixel Grounding}} \\ 
\cmidrule(lr){3-6} \cmidrule(lr){7-10} 
  &  & \textbf{BLEU} & \textbf{Rouge} & \textbf{BERT} & \textbf{GPT} & \textbf{AP@30} & \textbf{AP@50} & \textbf{AP@70} & \textbf{mIoU}  \\ 
\midrule
Qwen2.5-VL~\cite{qwen2.5-VL} & 72B  & 18.46 & 27.91 &  82.83 & 51.84 & 51.60 & 47.50 & 40.80 & 44.17 \\ 
InternVL3~\cite{chen2024internvl} & 78B & 19.89 & 27.87 & 82.05 & \textbf{71.26} & 5.10 & 3.90 & 2.90 & 4.17 \\ 
\midrule
LLaVA~\cite{liu2023visual} & 7B & 5.02 & 13.84 & 81.54 & 26.02 & 23.30 & 16.80 & 13.60 & 18.07 \\ 
GroundingGPT~\cite{li2024groundinggpt} & 7B & 0.01 & 5.90 & 80.82 & 28.46 & 13.50 & 12.30 & 10.40 & 11.95 \\ 
LISA~\cite{lai2024lisa} & 7B & 2.99 & 10.85 & 78.93 & 13.80 & 21.00 & 16.50 & 14.60 & 17.61 \\ 
Sa2VA~\cite{yuan2025sa2va} & 8B & 0.54 & 3.90 & 78.55 & 13.20 & 68.80 & 63.50 & 56.20 & 58.74 \\ 
\midrule
\textbf{SafePLUG (Ours)} & 7B & \textbf{34.54} & \textbf{40.15} & \textbf{86.09} & 65.13 & \textbf{74.30} & \textbf{68.10} &  \textbf{59.30} & \textbf{64.07} \\ 
\bottomrule
\end{tabular}
\label{tab:rqa+seg}
\end{table*}

Figure~\ref{fig3-2}(d) visualizes the spatial distribution of bounding boxes in the Region QA task. We cluster all bounding box coordinates and then randomly sample 150 boxes for visualization. The spatial density map shows that the annotated regions are distributed broadly across the frame, with a mild concentration in central and lower areas corresponding to vehicles and road surfaces. Each frame contains an average of 5.6 bounding boxes.
The object category distribution within all annotated bounding boxes is shown in Figure~\ref{fig3-2}(e). The majority of annotated objects are cars, followed by trucks and pedestrians, which together account for the vast majority of traffic entities in urban scenes.

Finally, Figure~\ref{fig3-2}(f) summarizes the distribution of accident types across all 220K QA pairs. 
We follow the accident category introduced in the DoTA~\cite{yao2022dota}. 
The dataset covers eight major accident types: (1) ST: collision with a vehicle that starts, stops, or is stationary; (2) AH: collision with a vehicle moving ahead or waiting; (3) LA: collision with a vehicle moving laterally in the same direction; (4) OC: collision with an oncoming vehicle; (5) TC: collision with a vehicle turning into or crossing a road; (6) VP: collision with a pedestrian; (7) VO: collision with an obstacle; and (8) OO: loss of control. 
Non-accident samples (NM) are also included to help models distinguish normal driving scenes from accident events.
The category distribution remains balanced, ensuring that the dataset provides sufficient diversity for robust and unbiased model learning.

\section{Experiments}

\subsection{Experimental Setting}

\subsubsection{Models and Training Configuration}

We employ the pretrained video and image encoders from LanguageBind~\cite{zhu2023languagebind}, which remain frozen during training to preserve their multimodal alignment. The projection layers used for mapping visual embeddings into the language space are initialized from Video-LLaVA~\cite{lin2023video}. For each video, we uniformly sample 8 frames as input to the video encoder to capture temporal context with balanced computational efficiency. The backbone language model is Vicuna-7B v1.5~\cite{chiang2023vicuna}, which serves as the main reasoning component for multimodal fusion and response generation.
For pixel-level segmentation, we adopt the SAM ViT-H~\cite{kirillov2023segment} model, where the pixel encoder is frozen and only the decoder is fine-tuned to learn safety-critical spatial segmentation. Training is conducted on 8$\times$A100 GPUs (80\,GB each). Additional training implementation details are provided in Appendix~\ref{sec:appendix-exp}.

\subsubsection{Evaluation Metrics}
For text-based tasks, we adopt BLEU-1~\cite{papineni2002bleu}, ROUGE-1~\cite{lin2004rouge}, and BERTScore F1~\cite{zhang2019bertscore} to measure linguistic similarity against ground-truth annotations. Following prior work~\cite{parikh2025roadsocial,xing2025echotraffic}, we further prompt GPT-3.5 to assess the generated responses in terms of consistency, reasonableness, and level of detail. More details of the GPT-3.5 prompting template are provided in Appendix~\ref{sec:appendix-exp}. For pixel-level grounding and temporal localization tasks, we report the Average Precision (AP) at thresholds of 30, 50, and 70 (denoted as AP@30, AP@50, and AP@70), together with mean Intersection over Union (mIoU). All metric scores are linearly scaled to a range of 0–100 for consistent comparison across tasks.

\begin{table*}[t]
\centering
\caption{Performance comparison on accident description and temporal localization. All metric scores range from 0 to 100, with the best performance highlighted in \textbf{bold}.}
\begin{tabular}{l |c| cccc | cccc}
\toprule
\multirow{2}{*}{\textbf{Model}} & \multirow{2}{*}{\textbf{Param.}} & \multicolumn{4}{c|}{\textbf{Accident Description}} & \multicolumn{4}{c}{\textbf{Temporal Localization}} \\ 
\cmidrule(lr){3-6} \cmidrule(lr){7-10} 
 &   & \textbf{BLEU} & \textbf{Rouge} & \textbf{BERT} & \textbf{GPT} & \textbf{AP@30} & \textbf{AP@50} & \textbf{AP@70} & \textbf{mIoU} \\ 
\midrule
Qwen2.5-VL~\cite{qwen2.5-VL} & 72B & 15.94 & 29.98 & 83.37 & 47.11 & 19.40 & 11.80 & 3.00 & 11.24 \\ 
InternVL3~\cite{chen2024internvl} & 78B & 1.98 & 8.38 & 80.11 & 19.20 & 1.40 & 0.60 & 0.00 & 2.44 \\ 
\midrule
Video-LLaVA~\cite{lin2023video}  & 7B & 3.98 & 17.11 &  81.78 & 19.44 & 44.00 & 17.80 & 3.00 & 25.93 \\ 
GroundingGPT~\cite{li2024groundinggpt}  & 7B & 3.66 & 13.74 & 81.28 & 19.12 & 3.00 & 0.20 & 0.00 & 2.85 \\ 
TimeChat~\cite{ren2024timechat}  & 7B & 0.63 & 9.78 & 80.93 & 17.75 & 1.60 & 0.00 & 0.00& 1.44\\ 
RoadSocial~\cite{parikh2025roadsocial}  & 7B  & 0.04 & 11.39 & 82.02 & 30.39 & 7.00 & 2.20 & 0.40 & 5.66 \\ 
\midrule
\textbf{SafePLUG (Ours)} & 7B & \textbf{30.29} & \textbf{38.31} & \textbf{85.49} & \textbf{66.47}  & \textbf{65.60} & \textbf{45.40} & \textbf{19.60} & \textbf{43.18} \\ 
\bottomrule
\end{tabular}
\label{tab:tl+ad}
\end{table*}

\subsection{Performance Evaluation}

We evaluate SafePLUG across four key tasks: region-level QA, pixel-level grounding, accident description, and temporal localization. All experiments are conducted on the test split of our proposed SafePLUG-Bench. Tables~\ref{tab:rqa+seg} and~\ref{tab:tl+ad} present quantitative comparisons against a diverse set of existing MLLMs. These baselines include both general-purpose vision-language models and specialized frameworks designed for spatiotemporal understanding. Given the substantial computational requirements of large multimodal models, all reported results are obtained from a single evaluation run using a fixed random seed to ensure reproducibility and fairness.

\subsubsection{Region-Level Question Answering}

For models that do not support region-specific visual prompts, we incorporate bounding box coordinates into the input prompt as additional textual descriptions. As shown in Table~\ref{tab:rqa+seg}, SafePLUG achieves the best overall performance across all textual metrics (BLEU, ROUGE, and BERTScore) and ranks competitively on GPT-based evaluation.
Compared with general-purpose vision–language models such as Qwen2.5-VL and LLaVA, SafePLUG yields absolute gains of over 15--20 points on BLEU and ROUGE, indicating stronger alignment between textual responses and reference region descriptions. These improvements stem from SafePLUG’s ability to directly attend to arbitrary-shaped visual prompts, which guide the language model to focus on semantically relevant subregions rather than relying on global scene context. Notably, SafePLUG attains a GPT score of 65.13, approaching the best 71.26 achieved by InternVL3 while using an order-of-magnitude smaller model (7B vs.\ 78B). These results suggest that incorporating explicit region-level visual grounding is more effective than simply scaling model size or relying solely on textual heuristics.

\subsubsection{Pixel Grounding}

Among all evaluated baselines, only LISA and Sa2VA natively support mask-level outputs. For fair comparison, we prompt other baseline models to output bounding box coordinates and convert them into masks using SAM. As shown in Table~\ref{tab:rqa+seg}, both LISA and Sa2VA exhibit limited generalization to complex accident scenes, achieving modest mean IoU values of 17.61 and 58.74, respectively. In contrast, SafePLUG achieves the highest performance across all metrics, with AP@50 = 68.10 and mIoU = 64.07, outperforming Sa2VA by over 5 points.

These improvements stem from SafePLUG’s design that tightly couples pixel-level reasoning with language understanding. By introducing the special \texttt{<SEG>} token into the LLM vocabulary and projecting its hidden representation to the SAM decoder input, SafePLUG enables the language model to produce segmentation-aware embeddings that align linguistic cues with spatial details. This explicit coupling allows SafePLUG to delineate fine-grained accident-related regions, such as overlapping vehicles and collision impact zones, which existing MLLMs often miss due to their coarse spatial grounding.

Furthermore, the consistent gains across all AP thresholds (AP@30–70) demonstrate that SafePLUG not only improves overall segmentation accuracy but also enhances robustness under stricter localization criteria. These results confirm that pixel-level understanding in traffic accident scenarios remains challenging even for advanced MLLMs, and that SafePLUG-Bench provides a valuable benchmark for fine-grained spatial grounding under realistic traffic conditions.

\subsubsection{Accident Description}

% In the accident description task, SafePLUG shows clear advantages in generating descriptions of traffic accidents. The superior performance stems from the design of our dataset, which emphasizes detailed causal and contextual annotations across a wide range of accident scenarios. In contrast, existing models often generate vague or overly generic outputs. These results further emphasize the complexity of traffic accident understanding. 

In the accident description task, SafePLUG demonstrates clear advantages in generating coherent and causally grounded narratives of traffic incidents. As shown in Table~\ref{tab:tl+ad}, SafePLUG surpasses all baselines across BLEU, ROUGE, BERTScore, and GPT-based metrics, achieving a GPT score of 66.47, which is more than double that of RoadSocial and significantly higher than large-scale models such as Qwen2.5-VL and InternVL3. The strong performance under both lexical (BLEU, ROUGE) and semantic (BERTScore, GPT) measures indicates that SafePLUG not only produces text closely aligned with reference annotations but also captures deeper causal and contextual semantics.

These improvements primarily stem from two design aspects of SafePLUG. First, our LoRA-based training pipeline employs a language-oriented adapter configuration that strengthens cross-modal alignment for causal narrative generation, enhancing coherence and factual grounding in accident descriptions. Second, the proposed SafePLUG-Bench provides richer causal and contextual supervision than prior datasets, incorporating diverse accident causes, object interactions, and environmental cues. This diversity enables SafePLUG to generalize beyond surface-level descriptions and generate detailed explanations that correctly attribute accident responsibility or identify precipitating maneuvers.
The consistent improvements across all evaluation metrics validate the role of SafePLUG-Bench as a challenging and comprehensive benchmark for measuring causal and contextual understanding in multimodal accident analysis.

\begin{figure}[!t]
\centering
\includegraphics[width=0.48\textwidth]{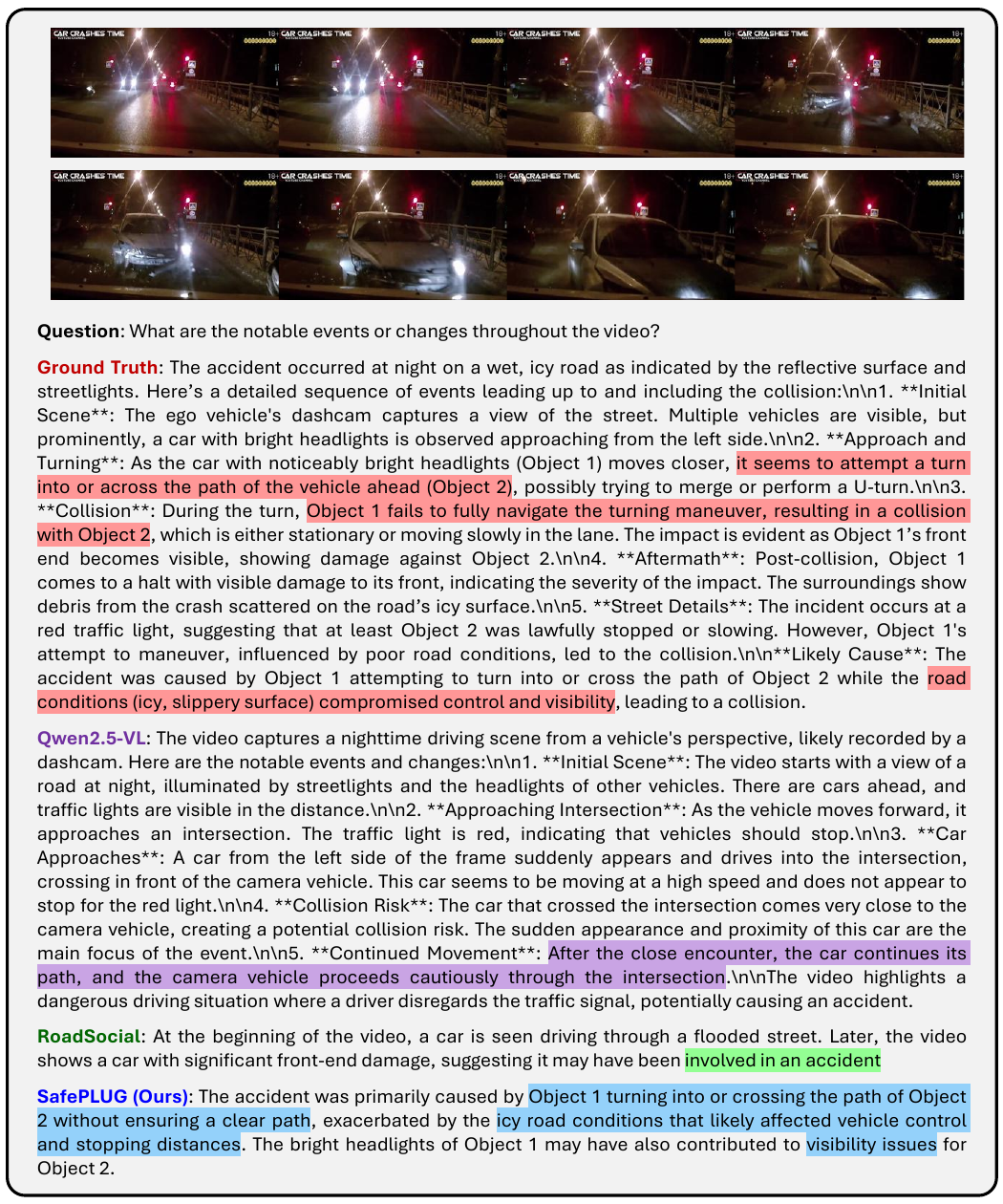}
\caption{Qualitative comparison of accident descriptions generated by different models. The highlighted colors indicate key causal phrases, agent actions, and event interpretations.}
\label{fig4}
\end{figure}

\subsubsection{Temporal Localization}

SafePLUG exhibits strong temporal localization capability, outperforming all baselines on SafePLUG-Bench by a considerable margin. As shown in Table~\ref{tab:tl+ad}, SafePLUG achieves an AP@50 of 45.40 and an mIoU of 43.18, substantially exceeding the best competing model, Video-LLaVA. These results confirm that SafePLUG can accurately identify the onset and offset of accident events, maintaining robustness even under stricter temporal thresholds (AP@70 = 19.60). The key factor behind this improvement is the use of number prompts, which provide effective temporal cues without modifying the model architecture. By embedding numerical indicators directly into video frames, the model learns to associate language expressions such as “when the collision occurred” with specific temporal indices. This alignment between temporal order and semantic content enables the language model to infer accident boundaries with higher precision and temporal coherence.

Although prior approaches such as GroundingGPT and TimeChat are designed for temporal grounding, they underperform on SafePLUG-Bench, likely due to limited exposure to complex traffic accident scenarios during training. In contrast, SafePLUG benefits from the fine-grained temporal annotations and diverse event patterns in SafePLUG-Bench, allowing it to generalize across heterogeneous accident types and accurately capture transitions between pre-accident, impact, and post-accident phases.

\subsection{Qualitative Analysis}

\begin{figure}[!t]
\centering
\includegraphics[width=0.48\textwidth]{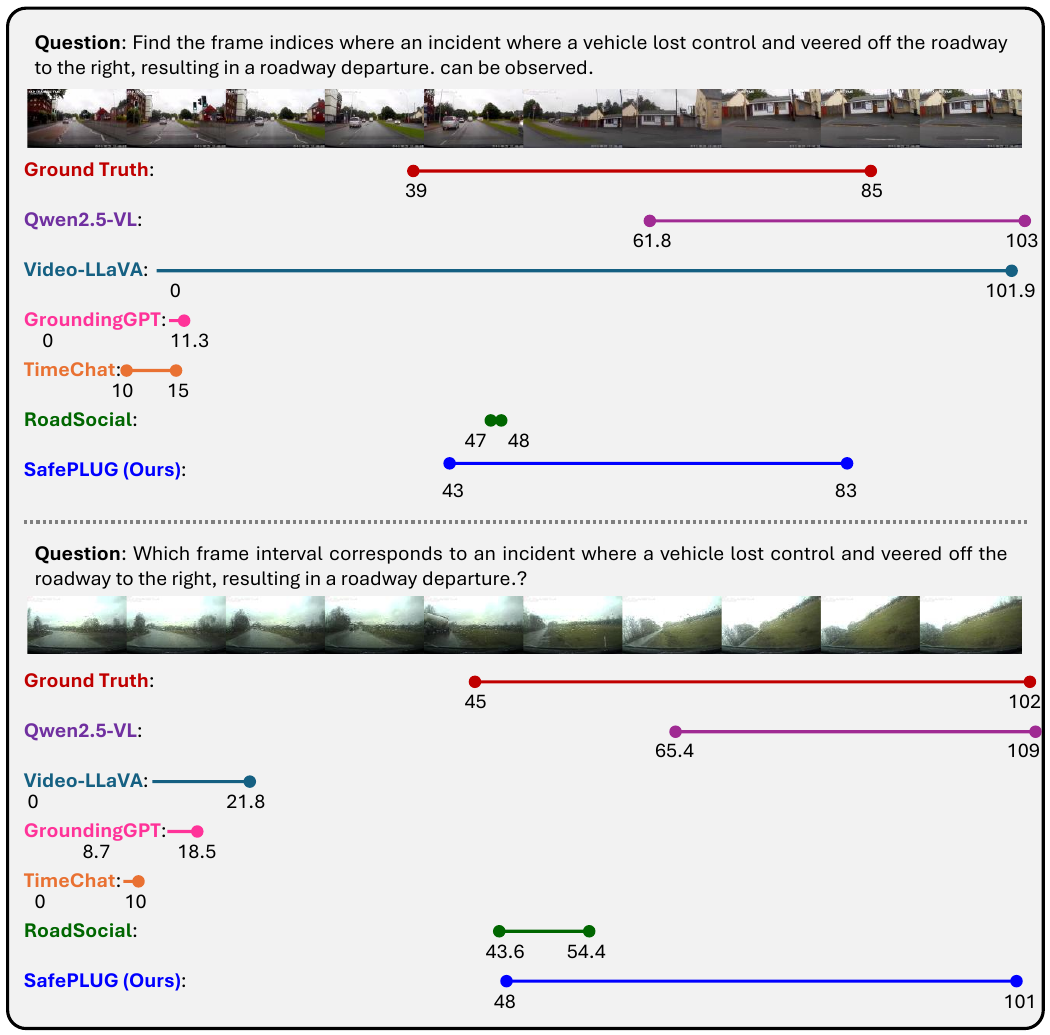}
\caption{Qualitative comparison of temporal localization results. Colored bars indicate the time spans from various models and the ground-truth annotation.}
\label{fig5}
\end{figure}

Beyond the quantitative results, we provide qualitative analysis in Figures~\ref{fig1} and \ref{fig4}--\ref{fig7} to visually illustrate the strengths of SafePLUG across all four tasks. These case studies demonstrate how SafePLUG achieves fine-grained and contextually consistent understanding of complex traffic accident scenes, accurately identifying involved agents, spatial regions, and temporal boundaries that baseline models often overlook or misinterpret. 

\subsubsection{Accident Description} 

Figure~\ref{fig4} presents a qualitative comparison of accident description outputs from baseline models and SafePLUG. We select Qwen2.5-VL and RoadSocial for comparison, as they achieve the highest GPT-based evaluation scores among all baselines. As shown in the figure, baseline models tend to produce scene-level summaries that omit critical causal relationships and misidentify the sequence of key events. For instance, Qwen2.5-VL focuses primarily on general road context and the near-miss interaction but fails to capture the actual collision and its underlying cause, while RoadSocial provides only a generic mention of vehicle damage without any causal interpretation.

In contrast, SafePLUG generates a concise yet causally grounded narrative that correctly identifies the roles of the involved agents and their interactions. Specifically, it infers that Object 1 attempted to turn into or cross the path of Object 2 without ensuring a clear path, and attributes the resulting collision to poor road conditions (icy, slippery surface) that compromised vehicle control and stopping distance. This explanation aligns closely with the ground-truth annotation, reflecting SafePLUG’s ability to integrate spatial cues, contextual reasoning, and linguistic coherence within a unified multimodal framework.

\subsubsection{Temporal Localization} 

As illustrated in Figure~\ref{fig5}, we visualize the accident phase boundaries predicted by SafePLUG and baseline models. Compared to TimeChat, GroundingGPT, and RoadSocial, which often produce overly short or misaligned temporal spans, SafePLUG consistently identifies both the onset and offset of accident events with high precision. Qwen2.5-VL provides closer predictions in some cases but still exhibits delayed or truncated intervals relative to the ground truth. This improvement stems from the integration of number prompts, which embed lightweight numerical indicators into video frames to provide effective temporal anchors. By learning to associate these indicators with semantic cues in the language query, SafePLUG can precisely map described events to their corresponding frame intervals.

\begin{figure}[!t]
\centering
\includegraphics[width=0.48\textwidth]{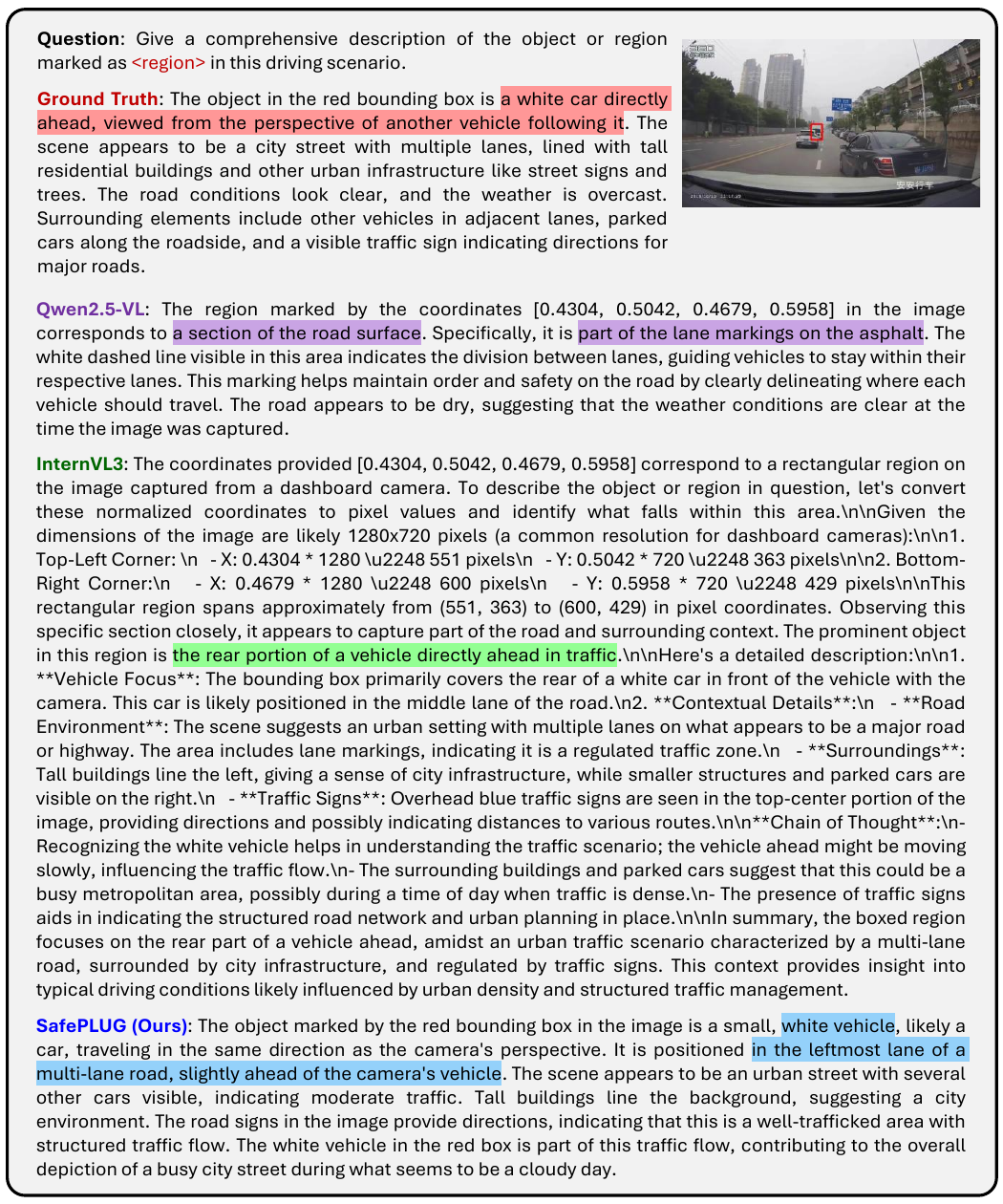}
\caption{Qualitative comparison of region-level question answering results. Each model is prompted to provide a detailed description of the region highlighted in the image. }
\label{fig6}
\end{figure}

\subsubsection{Region-Level Question Answering} 
Figure~\ref{fig6} shows qualitative examples of region QA. For fair comparison, baseline models such as Qwen2.5-VL and InternVL3 are provided with additional bounding box coordinates to specify the target region, whereas SafePLUG directly attends to the highlighted visual region via visual prompts.

Qwen2.5-VL generates a response focusing primarily on road surface markings, but fails to identify the key object (i.e., a white vehicle) within the region. InternVL3 provides a more detailed response using a chain-of-thought style, including contextual cues about traffic signs and road infrastructure. However, its output is overly verbose and occasionally redundant. In contrast, SafePLUG produces concise, spatially accurate, and semantically complete descriptions. It correctly identifies the object type, position, and relational context (e.g., a white car traveling ahead on a multi-lane urban road), aligning closely with the ground truth. This demonstrates the model’s capability to integrate localized visual cues with contextual reasoning to achieve fine-grained scene understanding.

\begin{figure}[!t]
\centering
\includegraphics[width=0.48\textwidth]{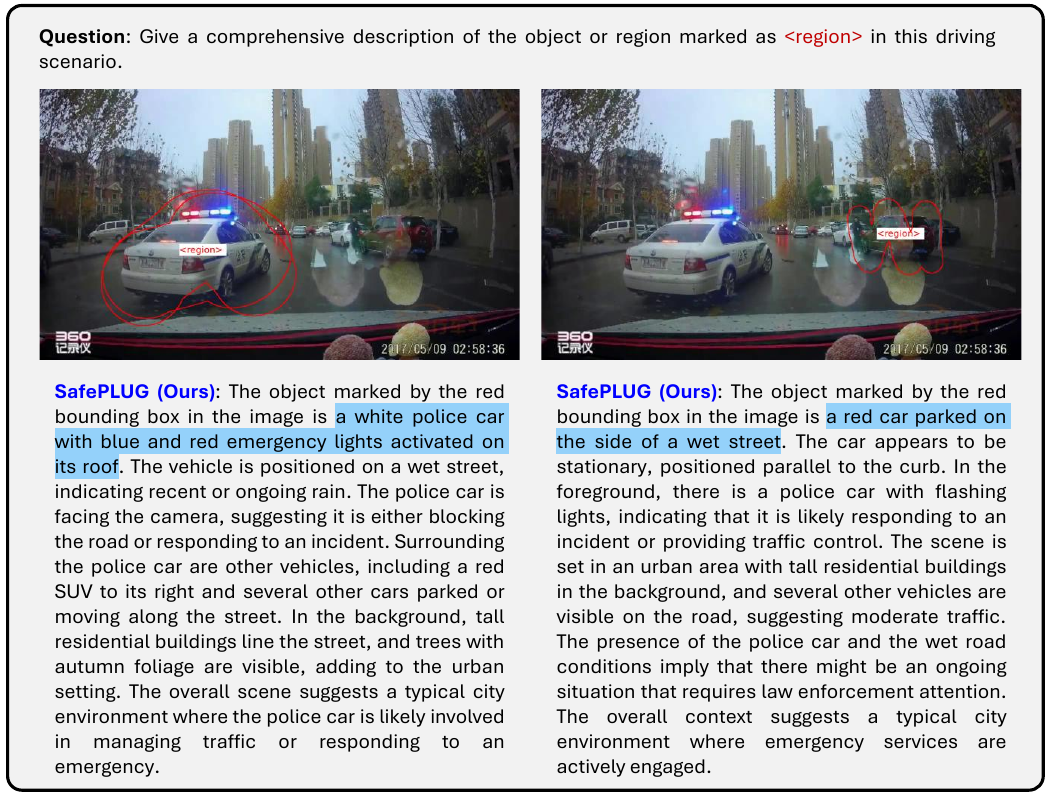}
\caption{Qualitative results demonstrating SafePLUG’s generalization ability across different visual prompt shapes.}
\label{fig6-2}
\end{figure}

To further assess the robustness of SafePLUG under diverse visual prompt shapes and positions, Figure~\ref{fig6-2} shows two cases where distinct free-form masks are used to mark different spatial regions within the same scene. Across both examples, SafePLUG accurately interprets each highlighted region and recognizes objects such as a police vehicle with flashing emergency lights and a red parked car on a street. These results indicate that SafePLUG generalizes effectively to arbitrary prompt shapes and spatial placements, confirming the flexibility and precision of its visual prompt encoder. 

\subsubsection{Pixel-Level Segmentation}

\begin{figure}[!t]
\centering
\includegraphics[width=0.48\textwidth]{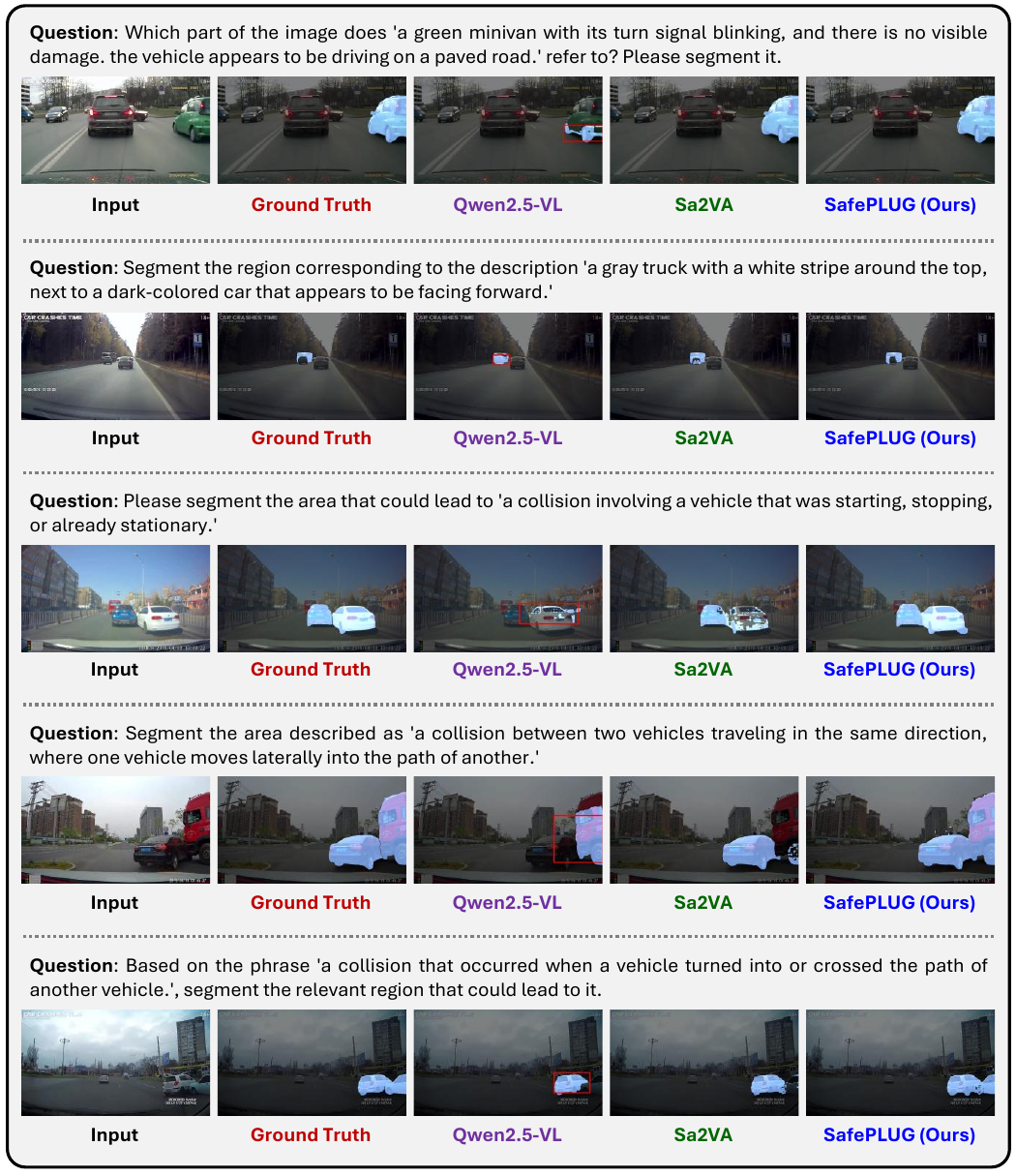}
\caption{Qualitative comparisons of pixel-level grounding results across multiple models. Each row presents a language-based segmentation query and the corresponding outputs from Qwen2.5-VL, Sa2VA, and SafePLUG, alongside ground-truth annotations. The red bounding boxes in Qwen2.5-VL outputs denote predicted regions, which are converted into segmentation masks using SAM.}
\label{fig7}
\end{figure}

Figure~\ref{fig7} presents qualitative comparisons of pixel-level segmentation results for SafePLUG, Qwen2.5-V, and Sa2VA across five representative scenarios. The segmentation queries range from simple object-level references (e.g., ``a green minivan”) to complex causal descriptions involving multiple agents (e.g., ``a collision between two vehicles traveling in the same direction”). Qwen2.5-VL generates bounding boxes that are post-processed into masks using SAM, whereas both Sa2VA and SafePLUG natively produce pixel-level mask generation.

Across all examples, SafePLUG yields segmentation masks that are spatially precise and semantically aligned with the described entities. It accurately delineates vehicle contours and captures subtle spatial relationships such as overlapping agents, even in cluttered or low-visibility scenes. By contrast, Sa2VA occasionally produces incomplete masks, while Qwen2.5-VL often highlights irrelevant background regions.

% These improvements arise from SafePLUG’s integration of the special \texttt{<SEG>} token, which bridges the LLM’s hidden representations with dense visual features extracted from the SAM-based pixel encoder. This design allows the model to translate high-level language cues into spatial attention, enabling fine-grained segmentation guided by semantic intent. Moreover, the diverse pixel-level annotations provided in \textsc{SafePLUG-Bench} expose the model to a wide range of accident contexts, thereby enhancing its robustness and generalization.

\subsection{Ablation Study}

We conduct ablation studies to assess the contribution of the multi-LoRA training strategy and key model components in SafePLUG. We report BLEU-1 scores for Region QA and Accident Description, and mIoU for pixel-level grounding and temporal localization, as summarized in Tables~\ref{tab:ablation_lora} and \ref{tab:module}.

\begin{table*}[t]
\centering
\caption{Effect of using different LoRA branches.}
\begin{tabular}{c|cc|cccc|c}
\toprule
\textbf{Configuration} & \shortstack{\textbf{Text}\\\textbf{LoRA}} & \shortstack{\textbf{Mask}\\\textbf{LoRA}} & \shortstack{\textbf{Region}\\\textbf{QA}} & \shortstack{\textbf{Pixel}\\\textbf{Grounding}} & \shortstack{\textbf{Accident}\\ \textbf{Description}} & \shortstack{\textbf{Temporal}\\ \textbf{Localization}}  & \textbf{Mean Score}  \\
 \midrule
(a) & \ding{55} & \ding{55} & 12.16 & 0.00 & 2.95 & 2.14 & 4.31 \\ 
(b) & \ding{51} & \ding{55} & \textbf{35.14} & 0.05 & \textbf{30.97} & 41.56 & 26.93 \\ 
(c) & \ding{55} & \ding{51} & 0.02 & \textbf{64.12} & 0.03 & 0.00 & 16.04 \\ 
(d) & \ding{51} & \ding{51} & 34.54 & 64.07 & 30.29 & \textbf{43.18} & \textbf{43.02} \\ 
\bottomrule
\end{tabular}
\label{tab:ablation_lora}
\end{table*}

\subsubsection{Effectiveness of Dual-LoRA Training}

To examine the effectiveness of the proposed dual-LoRA design, we conduct an ablation analysis in which each LoRA branch is selectively enabled or disabled during training and inference. As summarized in Table~\ref{tab:ablation_lora}, we compare four configurations: (a) a frozen backbone without any LoRA adaptation, (b) the text-oriented LoRA branch only, (c) the mask-oriented LoRA branch only, and (d) both LoRA branches jointly applied.

The results show the specialization of the two LoRA branches. Without any LoRA adaptation, the model fails to generalize across either textual or spatial modalities, indicating that the frozen backbone alone is insufficient to support downstream tasks. Enabling only the text-oriented LoRA substantially improves performance on language-based tasks such as region-level QA and accident description. This improvement demonstrates that the language-oriented adapters effectively enhance semantic understanding and cross-modal reasoning. However, since this configuration lacks the mask-oriented adaptation, its performance on pixel-level grounding still remains limited.

Conversely, activating only the mask-oriented LoRA yields the opposite trend: strong segmentation capability but severely degraded text reasoning. The model accurately localizes visual regions and produces high-quality binary masks, which confirms that the mask-oriented LoRA successfully refines dense spatial representations. Nonetheless, the absence of the text branch leads to weaker language grounding and insufficient semantic context, resulting in poor alignment between visual and textual modalities. This asymmetry highlights the necessity of having both LoRA branches to handle heterogeneous supervision signals.

When both branches are jointly loaded, SafePLUG achieves consistently strong results across all evaluation tasks. The combined configuration inherits the linguistic expressiveness of the text-oriented LoRA and the spatial precision of the mask-oriented LoRA, leading to the best overall performance. Importantly, the modular design allows each LoRA to specialize independently while remaining harmonized through the shared multimodal backbone. These findings validate that the dual-LoRA architecture enables SafePLUG to flexibly switch between textual reasoning and pixel-level understanding according to task requirements without retraining or compromising performance.

\begin{table*}[t]
\centering
\caption{Effect of the different modules. NP: Number Prompt, VP: Visual Prompt, PD: Pixel Decoder.}
\begin{tabular}{c|c|cccc|c}
\toprule
\textbf{Configuration} & \textbf{Model} & \shortstack{\textbf{Region}\\\textbf{QA}} & \shortstack{\textbf{Pixel}\\\textbf{Grounding}} & \shortstack{\textbf{Accident}\\ \textbf{Description}} & \shortstack{\textbf{Temporal}\\ \textbf{Localization}} & \textbf{Mean Score}  \\ 
\midrule
(a) & W/o NP & 34.89 & \textbf{64.18} & 30.06 & 28.33 & 39.37 \\ 
(b) & W/o VP & 18.75 & 63.93 & 30.89 & \textbf{42.11} & 38.92 \\ 
(c) & W/o PD & \textbf{35.32} & 20.46 & \textbf{31.08} & 41.55 & 32.10 \\ 
(d) & SafePLUG & 34.54 & 64.07 & 30.29 & 43.18 & \textbf{43.02}  \\ 
\bottomrule
\end{tabular}
\label{tab:module}
\end{table*}

\subsubsection{Effectiveness of Model Components}

We further conduct ablation experiments to assess the contribution of each key component within SafePLUG, including the number prompt, visual prompt, and pixel decoder. These components correspond to temporal grounding, region QA, and pixel-level segmentation, respectively.

As presented in Table~\ref{tab:module}, removing the number prompt (a) leads to a substantial degradation in temporal localization accuracy. Without explicit numeric cues embedded in video frames, the model struggles to associate semantic events with their corresponding temporal segments, often producing misaligned or truncated event intervals. This highlights the effectiveness of the number prompt that allows the model to implicitly learn frame-level ordering without requiring architectural modifications or additional supervision.

Removing the visual prompt (b) significantly reduces performance on region-level QA. In this variant, the model receives only the global image representation, losing its ability to focus on spatially localized areas of interest. Consequently, its responses become more generic and less contextually precise, demonstrating that visual prompts are essential for spatially grounded understanding. By guiding the model’s attention toward user-specified regions or object masks, the visual prompt mechanism enables SafePLUG to handle arbitrary-shaped regions and capture localized interactions critical to accident interpretation.

Eliminating the pixel decoder (c) yields a sharp drop in pixel-level grounding performance, even though textual and regional reasoning remain largely unaffected. This result confirms that dense segmentation cannot be achieved by the LLM alone, as it lacks the spatial resolution required for mask generation. The SAM-based pixel decoder serves as a crucial bridge between the LLM’s abstract representations and spatially detailed outputs, enabling accurate reconstruction of fine object boundaries and collision regions.

Finally, the complete model (d) achieves the highest mean performance across all tasks, indicating that each component contributes complementary benefits to SafePLUG’s overall capability. The number prompt enhances temporal precision, the visual prompt provides spatial selectivity, and the pixel decoder ensures pixel-level fidelity. Together, these modules enable SafePLUG to perform coherent multimodal reasoning that unifies spatial, temporal, and semantic understanding within a single framework.

\section{Conclusion}
In this work, we propose SafePLUG, a novel framework that empowers multimodal large language models with both pixel-level understanding and temporal grounding capabilities for comprehensive traffic accident understanding. By integrating visual prompts for region-aware reasoning, number prompts for implicit temporal cues, and SAM for fine-grained segmentation, SafePLUG enables detailed spatial and temporal analysis across diverse accident scenarios. We further construct a large-scale benchmark dataset, SafePLUG-Bench, to support region QA, pixel-level grounding, accident description, and temporal localization. Extensive experiments demonstrate that SafePLUG outperforms strong baselines across all tasks while maintaining a lightweight architecture. Ablation studies confirm the effectiveness of our dual-LoRA training and modular design. In future work, we plan to extend SafePLUG to support long-range video reasoning, richer multimodal contexts such as audio and sensor data, and real-time deployment for traffic monitoring.

% if have a single appendix:
%\appendix[Proof of the Zonklar Equations]
% or
%\appendix  % for no appendix heading
% do not use \section anymore after \appendix, only \section*
% is possibly needed

% use appendices with more than one appendix
% then use \section to start each appendix
% you must declare a \section before using any
% \subsection or using \label (\appendices by itself
% starts a section numbered zero.)
%

\appendices

\section{Additional Details on Experiment Settings}
\label{sec:appendix-exp}

We conducted all experiments on a machine equipped with 8 NVIDIA A100 GPUs, each with 80GB of memory, running Ubuntu 22.04. The core software environment consists of PyTorch, DeepSpeed, and Hugging Face Transformers. To ensure reproducibility, we fixed all sources of randomness by setting a unified random seed across Python, NumPy, and PyTorch, and enforcing deterministic behavior in cuDNN.

For visual encoding, we follow Video-LLaVA~\cite{lin2023video} by adopting CLIP-L/14~\cite{radford2021learning} as the image encoder and the video encoder from LanguageBind~\cite{zhu2023languagebind}. The extracted visual features are projected into the language model space using a two-layer MLP with a GELU activation function~\cite{hendrycks2016gaussian}. For the mask-oriented training, we follow prior works~\cite{lai2024lisa,yan2024visa} by applying a weighted combination of BCE loss and DICE loss to jointly enhance mask coverage and boundary precision. Specifically, the BCE loss weight $\lambda_{\text{BCE}}$ is set to 2.0, and the DICE loss weight $\lambda_{\text{DICE}}$ is set to 0.5. We also include a text generation objective with a cross-entropy loss weighted by 1.0. We train the language-oriented LoRA branch for 5 epochs with an initial learning rate of 0.0001 and the mask-oriented LoRA branch for 20 epochs with an initial learning rate of 0.0003 on our proposed dataset.
A relatively higher LoRA rank ($r=16$) is assigned to the language-oriented branch to provide greater representational flexibility for capturing complex linguistic dependencies. In contrast, the mask-oriented branch employs a smaller rank ($r=8$) to ensure stable optimization and mitigate overfitting to local spatial patterns.

In addition to standard metrics such as BLEU, ROUGE, and BERTScore, we also use GPT-3.5 as an LLM evaluator to assess the quality of generated text responses. Following prior work~\cite{parikh2025roadsocial,xing2025echotraffic}, we prompt GPT-3.5 with three components: the input question, the reference answer, and the model-generated response. The GPT model is instructed to rate the output based on its reasonableness, level of detail, and consistency with the reference answer. The score ranges from 0 to 100, with higher values indicating better alignment with the reference. The prompt used for GPT-3.5 evaluation is shown in Table~\ref{tab:gpt}.

\begin{table*}[!t]
\centering
\caption{Prompt template used for GPT-3.5-based evaluation. The model receives the question, reference answer, and predicted answer, and returns an integer score along with a brief explanation based on reasonableness, detail, and consistency.}
\begin{tcolorbox}[halign=left]

\textbf{System Message}

You are a helpful and precise assistant for checking the quality of the answer.\\[6pt]

\textbf{User Message}

Evaluate the following question-answer pair:

Question: \texttt{<QUESTION>}

Correct Answer: \texttt{<REFERENCE>}

Predicted Answer: \texttt{<ANSWER>}

Rate the Predicted Answer based on the Correct Answer on a scale from 0 to 100, with higher scores indicating that the Predicted Answer is closer to the Correct Answer. Your rating should be accurate to single digits like 60, 33, 87, etc.

Your rating should consider the \textbf{reasonableness}, \textbf{detail}, and \textbf{consistency}.
Please generate the response in the form of a Python dictionary string with keys ``score", where its value is in INTEGER, not STRING, and ``explanation" giving short and concise reasoning behind the score.

For example, your response should look like this: \{``score": 38, ``explanation": ``..."\}
\end{tcolorbox}
\label{tab:gpt}
\end{table*}

\section{Additional Details on Dataset Construction}
\label{sec:appendix_dataset}

To construct high-quality multimodal QA pairs for region-level and pixel-level understanding, we design structured prompting templates and use large vision-language models for automated annotation, followed by manual verification. The prompting templates used for InternVL3-78B~\cite{chen2024internvl}, Qwen2.5-VL-72B~\cite{qwen2.5-VL}, and Qwen2.5-72B~\cite{qwen2.5} are illustrated in Table~\ref{tab:internvl+qwen}. These templates guide region description generation by presenting either the full image with overlaid bounding boxes or cropped regions to the models, and judge the semantic consistency between the two generated descriptions.

To construct QA pairs in our dataset, we design question templates for four tasks: Region QA, Pixel-level Grounding QA, Accident Description, and Temporal Localization. The full set of templates is provided in Tables~\ref{tab:q1}--\ref{tab:q4}.

To ensure annotation quality, we conduct human review for both accident descriptions and segmentation masks. For accident descriptions, we sample a subset where two experts examine whether the generated text contains hallucinated objects or events, conflicts with the annotated accident cause, or lacks coherent causal logic. During this process, we refine and finalize the prompt format to ensure accurate and consistent annotation across the entire dataset. For SAM-generated segmentation masks, six experts check and remove masks that are poorly aligned with the target region, overly coarse, or fragmented.

\begin{table*}[!t]
\centering
\caption{Prompting templates used for region description generation and consistency checking.}
\begin{tcolorbox}[halign=left]

\textbf{InternVL3-78B Prompt}

Please describe the object marked by the red bounding box in this image in detail. Additionally, explain the relationship between the boxed object and other elements within the overall context of the scene.

Use natural and complete English sentences and write a single, coherent paragraph.\\[6pt]

\textbf{Qwen2.5-VL-72B Prompt}

Please describe the object shown in the image. Focus on its appearance, type (e.g., vehicle, pedestrian), color, and any visible attributes such as damage or direction. 

Use a complete sentence.\\[6pt]

\textbf{Qwen2.5-72B Prompt}

Evaluate the following two descriptions and judge whether they refer to the same object and provide consistent semantic information.

Description 1: \texttt{<DESCRIPTION1>}

Description 2: \texttt{<DESCRIPTION2>}

Respond with ``Yes'' if they are consistent, or ``No'' if they describe different objects or contain conflicting information.

\end{tcolorbox}
\label{tab:internvl+qwen}
\end{table*}

\begin{table*}[!t]
\centering
\caption{Instruction templates used for constructing region QA prompts. Each template guides the model to describe or analyze the visual content within a specified region denoted as \texttt{<region>}.}
\begin{tcolorbox} 
    \centering
\begin{itemize}
    \item ``Please provide a detailed description of the content in \texttt{<region>} within the current traffic scene."
    \item ``Please describe the object shown in \texttt{<region>}."
    \item ``What is happening in \texttt{<region>}? Describe it in the context of surrounding road elements."
    \item ``What can be observed about the object in \texttt{<region>}?"
    \item ``Describe what you observe in \texttt{<region>}, considering the traffic environment."
    \item ``Can you explain the visual content of \texttt{<region>} and its role in the road context?"
    \item ``Give a comprehensive description of the object or region marked as \texttt{<region>} in this driving scenario."
    \item ``What does the object in \texttt{<region>} look like?"
    \item ``What can be seen in \texttt{<region>}?"
    \item ``Summarize the visual appearance of the object located in \texttt{<region>}."
    \item ``I’m interested in what’s inside \texttt{<region>}. Could you provide a detailed account?"
    \item ``Can you elaborate on the content shown in \texttt{<region>}?"
    \item ``What information does \texttt{<region>} convey visually? Please describe it with respect to the current driving situation."
    \item ``Provide an in-depth description of \texttt{<region>} and how it fits into the broader driving context."
    \item ``Analyze the scene shown in \texttt{<region>} and explain its significance in this traffic scenario."
    \item ``Describe \texttt{<region>} as if you are explaining its contents to a driver navigating the road."
    \item ``What is visually represented in \texttt{<region>}? Consider how it may affect traffic behavior."
    \item ``Please describe the region \texttt{<region>} and mention any notable interactions it may involve."
    \item ``Describe the main object within \texttt{<region>} in the context of the scene."
    \item ``Give a clear description of what is shown in \texttt{<region>} as an object."
    \item ``Share your observation of the object highlighted in \texttt{<region>}."
\end{itemize}
\end{tcolorbox}
\label{tab:q1}
\end{table*}

\begin{table*}[!t]
\centering
\caption{Instruction templates used for pixel-level grounding QA. Each template guides the model to segment the region or object referred to by a natural language description denoted as \texttt{<description>}.}
\begin{tcolorbox} 
    \centering
\begin{itemize}

    \item ``Segment the object referred to as `\texttt{<description>}'."
    \item ``Which region corresponds to the phrase `\texttt{<description>}'? Please segment it."
    \item ``Segment the object described as `\texttt{<description>}'."
    \item ``Can you find and segment the object that is referred to as `\texttt{<description>}'?"
    \item ``Please segment the object mentioned in the phrase `\texttt{<description>}'."
    \item ``Segment the region corresponding to the description `\texttt{<description>}'."
    \item ``Given the description `\texttt{<description>}', which area should be segmented?"
    \item ``Segment the object indicated by `\texttt{<description>}'."
    \item ``What does the phrase `\texttt{<description>}' refer to in this image? Segment it."
    \item ``Find the object described as `\texttt{<description>}', and generate its segmentation."
    \item ``Based on the phrase `\texttt{<description>}', segment the relevant region."
    \item ``Determine the segmentation mask corresponding to `\texttt{<description>}'."
    \item ``Draw the segmentation of the entity mentioned in `\texttt{<description>}'."
    \item ``Which part of the image does `\texttt{<description>}' refer to? Please segment it."
    \item ``Segment the most likely object corresponding to `\texttt{<description>}'."
    \item ``Use the phrase `\texttt{<description>}' to segment the object."
    \item ``With the referring expression `\texttt{<description>}', produce the corresponding segmentation."
    \item ``Segment the part of the image that is being described as `\texttt{<description>}'."
    \item ``Which instance is being referred to as `\texttt{<description>}'? Please segment it."
    \item ``From the instruction `\texttt{<description>}', determine and segment the correct object."
    \item ``Segment the area that could lead to `\texttt{<description>}'."
    \item ``Which region could lead to `\texttt{<description>}'? Please segment it."
    \item ``Segment the area described as `\texttt{<description>}'."
    \item ``Can you find and segment the area that could lead to `\texttt{<description>}'?"
    \item ``Please segment the area that could lead to `\texttt{<description>}'."
    \item ``Based on the phrase `\texttt{<description>}', segment the relevant region that could lead to it."
\end{itemize}
\end{tcolorbox}
\label{tab:q2}
\end{table*}

\begin{table*}[!t]
\centering
\caption{Instruction templates used for the accident description task.}
\begin{tcolorbox} 
    \centering
\begin{itemize}
    \item ``Please describe what is happening in this driving video."
    \item ``Give a summary of the events unfolding in the scene."
    \item ``What can be observed throughout this traffic video?"
    \item ``Generate a description of the overall situation shown in the video."
    \item ``Briefly explain the sequence of events in this driving scenario."
    \item ``What is taking place on the road in this video?"
    \item ``Provide a natural language description of the traffic scene."
    \item ``Describe the key activities or motions occurring in this driving footage."
    \item ``Write a caption that summarizes the dynamic visual content."
    \item ``What are the notable events or changes throughout the video?"
    \item ``Based on the video, what is the main situation being presented?"
    \item ``Summarize the traffic-related activity depicted in the video."
    \item ``Give a general narrative of what is seen in this video segment."
    \item ``Provide a coherent and fluent description of the scene evolution."
    \item ``Describe how the situation unfolds in the driving environment."
    \item ``What is the traffic context or situation illustrated in the video?"
    \item ``How would you explain the scene to someone not watching the video?"
    \item ``Generate a description of what happens from start to end."
\end{itemize}
\end{tcolorbox}
\label{tab:q3}
\end{table*}

\begin{table*}[!t]
\centering
\caption{Instruction templates used for the temporal localization task. Each template guides the model to identify the frame interval during which a described event, denoted by \texttt{<description>}, occurs in the video sequence.}
\begin{tcolorbox} 
    \centering
\begin{itemize}
    \item ``During which frames can we see \texttt{<description>}?"
    \item ``In which frames does \texttt{<description>} appear?"
    \item ``Identify the frames where \texttt{<description>} is visible."
    \item ``From which frame to which frame does \texttt{<description>} occur?"
    \item ``Can you tell me the frame range where \texttt{<description>} is happening?"
    \item ``Which frames contain the event: \texttt{<description>}?"
    \item ``Around which frames does \texttt{<description>} take place?"
    \item ``Find the frame indices where \texttt{<description>} can be observed."
    \item ``What is the frame duration of \texttt{<description>} in the sequence?"
    \item ``Which frame interval corresponds to \texttt{<description>}?"
    \item ``Mark the frames during which \texttt{<description>} is ongoing."
    \item ``During what frames can one observe \texttt{<description>} occurring?"
\end{itemize}
\end{tcolorbox}
\label{tab:q4}
\end{table*}

% use section* for acknowledgment
\section*{Acknowledgment}

This work was supported by the Center for Connected and Automated Transportation (CCAT), the USDOT Region 5 University Transportation Center funded by the U.S. Department of Transportation, Award \#69A3552348305. The contents of this paper reflect the views of the authors, who are responsible for the facts and the accuracy of the data presented herein, and do not necessarily reflect the official views or policies of the sponsoring organization.

% The authors would like to thank...

% Can use something like this to put references on a page
% by themselves when using endfloat and the captionsoff option.
\ifCLASSOPTIONcaptionsoff
  \newpage
\fi

% trigger a \newpage just before the given reference
% number - used to balance the columns on the last page
% adjust value as needed - may need to be readjusted if
% the document is modified later
%\IEEEtriggeratref{8}
% The "triggered" command can be changed if desired:
%\IEEEtriggercmd{\enlargethispage{-5in}}

% references section

% can use a bibliography generated by BibTeX as a .bbl file
% BibTeX documentation can be easily obtained at:
% http://mirror.ctan.org/biblio/bibtex/contrib/doc/
% The IEEEtran BibTeX style support page is at:
% http://www.michaelshell.org/tex/ieeetran/bibtex/
\bibliographystyle{IEEEtran}
% argument is your BibTeX string definitions and bibliography database(s)
\bibliography{IEEEabrv,main}
\end{document}